%% file: main.tex
\documentclass[preprint]{article}

 \usepackage{neurips_2026}

\usepackage[utf8]{inputenc} 
\usepackage[T1]{fontenc}    
\usepackage{hyperref}       
\usepackage{url}            
\usepackage{booktabs}       
\usepackage{amsfonts}       
\usepackage{nicefrac}       
\usepackage{microtype}      
\usepackage{xcolor}         

\usepackage{wrapfig}
\usepackage{amsmath}
\usepackage{amssymb}
\usepackage{booktabs}
\usepackage{tabularx}
\usepackage{colortbl}
\usepackage{pifont}
\usepackage{epigraph}
\usepackage{subcaption}
\usepackage{rotating}
\usepackage{multirow}
\usepackage{soul}
\usepackage{graphicx}
\usepackage{caption}
\usepackage{xcolor} 
\usepackage{tikz}

\usepackage{tcolorbox}
\usepackage[normalem]{ulem} 
\usepackage{calc} 

\definecolor{lightblue}{RGB}{173,216,230}
\definecolor{codegreen}{rgb}{0,0.6,0}
\definecolor{neuripspurple}{RGB}{198,160,231}

\newcommand{\xmark}{\textcolor{red!70!black}{\ding{55}}}
\let\oldcheckmark\checkmark
\renewcommand{\checkmark}{\textcolor{green!60!black}{\ensuremath{\oldcheckmark}}}

\newtheorem{proposition}{Proposition}

\usepackage{orcidlink}

\newcommand{\modelname}{\textsc{UniEgo}}
\newcommand{\stageII}{Selective Proxy Distillation}
\newcommand{\stageIIabbr}{SPD}

\title{\modelname: Proxies as Mediators for Unified Egocentric Video Representation Learning}

\author{%
  Wenhao Chi, Arkaprava Sinha, Dominick Reilly, Hieu Le, Srijan Das
  \\
  University of North Carolina at Charlotte
}

\begin{document}
\maketitle

\begin{abstract}

Egocentric video understanding is inherently limited by the narrow perspective of wearable cameras: a single viewpoint, a single modality, a single model cannot capture the full richness of human action. We argue that a truly expressive egocentric representation must subsume complementary knowledge across viewpoints, modalities, and foundation model representations, yet remain deployable from egocentric video alone. To this end, we introduce a hierarchical multi-teacher distillation framework that produces \textbf{\modelname}, a unified egocentric encoder trained with nine teachers spanning ego-exo viewpoints, RGB, depth, and skeleton modalities, and four foundation models. Rather than distilling directly from heterogeneous teachers whose incompatible architectures and feature geometries induce conflicting gradients, our framework interposes a layer of representation-specific \textit{Proxy} models that translate diverse teacher knowledge into a homogeneous egocentric space. A second distillation stage, \textbf{\stageII~(\stageIIabbr)}, then adaptively selects, for each training sample, the subset of proxies that are both correct and confident, distilling exclusively from reliable supervision and suppressing erroneous signals. \stageIIabbr~is further stabilized by initializing \modelname~as a learned convex combination of proxy parameters, placing the unified model in a well-conditioned region of the loss landscape before distillation begins. \modelname~achieves state-of-the-art performance across three egocentric video understanding tasks - action recognition, video retrieval, and action segmentation on three challenging ego-exo benchmarks, outperforming naive multi-teacher distillation baselines and demonstrating that structured, proxy-mediated knowledge transfer yields richer and more discriminative egocentric representations. We release code and models at
\href{https://github.com/Wenhao-Chi/UNIEGO}{\textcolor{blue!70!black}{\texttt{https://github.com/Wenhao-Chi/UNIEGO}}}.
\end{abstract}

\input{sec/1_introduction}
\input{sec/2_related_works}

\input{sec/3_method}
\input{sec/4_experiments}

\input{sec/5_conclusion}

\bibliographystyle{plain}
\bibliography{ref,dominick_ref}

\newpage
\appendix
\input{sec/6_supplementary}

\end{document}

%% file: sec/1_introduction.tex
\section{Introduction}
\label{sec:intro}

Understanding human actions from egocentric video is a fundamental challenge in visual perception, with broad applications spanning augmented reality, assistive robotics, and procedural activity analysis~\cite{ego_exo_survey,Damen2018EPICKITCHENS,sigurdsson2018charades-ego}. Yet, learning a truly expressive egocentric representation from a single model remains elusive. Wearable cameras impose a narrow field of view and suffer from persistent self-occlusions, obscuring the actor's body and surrounding scene context~\cite{grauman2022ego4d}. Complementary modalities such as depth and skeleton which encode the geometric structure of human motion are discarded entirely~\cite{stgcn, videopose3d, OpenPose}. Moreover, the rich, diverse representational knowledge encapsulated within large-scale foundation models (FMs) remains untapped~\cite{zhai2023siglipv1, dino, yang2024depthanythingv2, clip_representation}. Consequently, a standalone egocentric model is fundamentally limited.
We argue that overcoming these limitations requires a \textit{unified egocentric representation}, a single, comprehensive embedding that subsumes complementary knowledge across modalities (RGB, depth, skeleton), viewpoints (egocentric and exocentric), and diverse representations from heterogeneous foundation models, yet operates solely on egocentric video at inference. Such a representation would unlock richer action understanding across a broad spectrum of downstream egocentric tasks.

These diverse perceptual signals spanning viewpoints, modalities, and FM representations naturally suggest a multi-teacher knowledge distillation framework, a paradigm explored extensively in concurrent work~\cite{ranzinger2024radio, shang2024theia, wimmer2026anyup, ticon}. However, existing methods distill homogeneous representations into a single student, sidestepping the deeper structural challenges that arise in the egocentric setting. First, teachers here are fundamentally heterogeneous: skeleton-based models operate over graph-structured neural architectures incompatible with video encoders~\cite{das2020vpn, vpn++}, while exocentric RGB teachers encode scene geometry from an entirely different viewpoint, introducing a substantial ``\textit{representational gap}"~\cite{luo2025viewpoint, li2021_egoexo-transfer}. Second, naïvely forcing an egocentric student to reconcile these incompatible feature spaces simultaneously leads to ``\textit{conflicting gradient signals}" and degrades optimization~\cite{conflict_adverse_GD, PCGRAD, shi2023recon}. Thus, the ego student is asked not merely to learn, but to simultaneously bridge modality gaps, close viewpoint gaps, and absorb diverse representational priors, an ill-posed objective for a single distillation stage. This motivates a more principled approach to unified egocentric representation learning.

\begin{figure}[h]
  \centering
  \includegraphics[width=\linewidth]{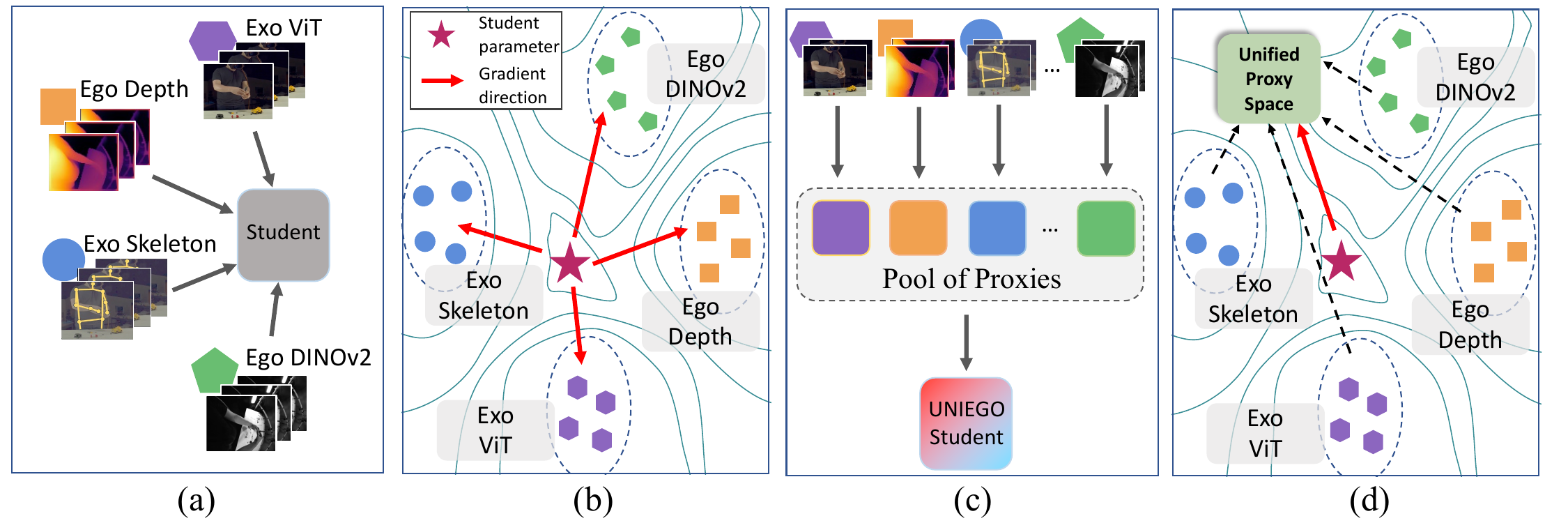}
  \caption{\textbf{(a)} Naive multi-teacher distillation with heterogeneous teachers for learning unified egocentric representations results in representational gaps and conflicting gradients, as illustrated in \textbf{(b)}. \textbf{(c)} In contrast, our proposed \textbf{\modelname}~adopts \textit{a hierarchical distillation framework} that mitigates these limitations through proxy-mediated learning, as shown in \textbf{(d)}. Black dashed arrows illustrate the effects of this framework, shifting teachers into a unified representation space.}
  \label{fig:teaser}
\end{figure}

We address these challenges by introducing a hierarchical multi-teacher distillation framework that consolidates diverse perceptual signals across viewpoints, modalities, and FM representations into \textbf{\modelname}, \textit{a single unified egocentric encoder} (see Figure~\ref{fig:teaser}). Rather than distilling directly from heterogeneous teachers, our framework first translates each teacher's knowledge into a student-compatible space through a set of representation-specific \textit{Proxy} models. Each proxy shares the architecture of \modelname~and operates on egocentric video. Therefore, the proxy learning diminishes the representational gap induced by heterogeneous teacher architectures and viewpoints by converting incompatible feature geometries into a homogeneous egocentric embedding space. 

Further, the first distillation level naturally exposes the reliability of each supervision signal on a per-instance basis: a proxy that cannot correctly classify a given sample carries no trustworthy knowledge to transfer. Our proposed hierarchical framework exploits this signal through \stageII~(\stageIIabbr), which selects, for each training sample, the subset of proxies whose predictions are both correct and confident, distilling exclusively from this reliable subset and suppressing erroneous supervision entirely. To further stabilize optimization, \stageIIabbr~is initialized via a learned convex combination of proxy parameters, placing \modelname~in a flat, well-conditioned region of the loss landscape prior to distillation. Together, proxy merging and proxy selection collectively mitigate the conflicting gradient problem inherent to naive multi-teacher distillation.

The outcome of our hierarchical distillation framework is \modelname, a unified egocentric encoder trained with 9 teachers spanning ego-exo viewpoints, RGB, depth, and skeleton modalities, and four FMs. \modelname~outperforms naive multi-teacher distillation baselines across three egocentric video understanding tasks - action recognition, video retrieval, and action segmentation on three challenging ego-exo benchmarks~\cite{Sener2022Assembly101AL, egoexofitness, EgoExo4D}. Moreover, \modelname~generalizes across video backbone architectures, including compact models with as few as 22M parameters. We summarize our contributions as follows:

\begin{itemize}
    \item We introduce \textbf{\modelname}, a unified egocentric encoder trained via a novel  hierarchical distillation framework with \textit{nine} teachers across \textit{ego-exo} viewpoints, \textit{three} modalities, and \textit{four} foundation models, using representation-specific proxies as structured mediators.

    \item We propose three tightly coupled components: \textbf{(i)} \textit{Proxy Learning}, which converts heterogeneous, multi-modal teacher supervision into a pool of architecturally homogeneous egocentric proxies, bridging the modality and viewpoint gap; \textbf{(ii)} \textit{{\stageII~(\stageIIabbr)}}, a sample-wise selective distillation mechanism that dynamically routes supervision from the most reliable proxies, mitigating conflicting gradients across heterogeneous teachers; and \textbf{(iii)} \textit{Proxy Merging}, a learned convex initialization of \modelname~that places the unified model in a well-conditioned region of the loss landscape, stabilizing \stageIIabbr's optimization.

    \item \modelname~achieves \textbf{state-of-the-art performance} across three egocentric video understanding tasks on three challenging benchmarks, demonstrating that hierarchical proxy-mediated distillation yields richer, more discriminative egocentric representations than direct multi-teacher supervision.
\end{itemize}

%% file: sec/2_related_works.tex
\section{Related Work}
\label{sec:relatedw}

\subsection{Egocentric Representation Learning}
Over the past few years, egocentric representation learning has become a central problem in video understanding. Early works, such as EgoVLP~\cite{kevin2022egovlp} and LaViLa~\cite{lavila}, focused on learning egocentric representations from egocentric videos alone~\cite{Damen2018EPICKITCHENS, grauman2022ego4d, sigurdsson2018charades-ego, egovlpv2}. This is challenging: egocentric cameras move with the person, hands and objects frequently occlude one another, and the same action can look very different depending on the wearer and environment. More recent works augment egocentric representation learning with signals beyond the raw ego video, typically leveraging synchronized exocentric viewpoints~\cite{li2021_egoexo-transfer, xue2023_egoexo-ae2, quattrocchi2024_synch-all-you-need, EMBED, luo2025viewpoint} or additional modalities~\cite{gong2023mmgego4d, xu2025egodtm, tan2023egodistill, radevski2023multimodal}. For example, ViewpointRosetta~\cite{luo2025viewpoint} uses diffusion models to learn a mapping between egocentric and exocentric representation spaces, EgoDTM~\cite{xu2025egodtm} learns 3D-aware egocentric representations through distillation from a depth-modality trained teacher. These methods typically exploit a single specific auxiliary signal (viewpoint/modality). In contrast our work aims to consolidate many heterogeneous teachers, spanning auxiliary viewpoints and modalities, into a single egocentric encoder.

\subsection{Multi-teacher knowledge distillation}
Knowledge distillation was originally introduced as a way to compress the knowledge of a large model, or an ensemble of models, into a single deployable student~\cite{knowledge_distillation_hinton2015}. This idea has since been extended to multi-teacher distillation, where a student learns from several teachers rather than a single teacher.
For example, approaches like AMTML-KD~\cite{liu2021adaptive} and CA-MKD~\cite{zhang2022confidence} adaptively distill from an ensemble of teachers operating on a single input modality, learning instance-level importance weights for each teacher. More recent work has considered a more heterogeneous form of multi-teacher distillation, where teachers are differentiated by their architecture or input modalities. 
For example, AM-RADIO~\cite{ranzinger2024radio} and Theia~\cite{shang2024theia} both distill multiple vision foundation models with different architectures and pretraining objectives into a single universal model. The key differences between these approaches and {\modelname} are that: (1) we learn egocentric representation learning that has not been explored by existing works, (2) we leverage a more diverse set of teachers spanning various architectures, modalities, and viewpoints; and (3) we selectively choose which teachers to distill from rather than treating all teacher as equally useful.

%% file: sec/3_method.tex
\begin{figure}[h]
  \centering
  \includegraphics[height=4.85cm, width=14cm]{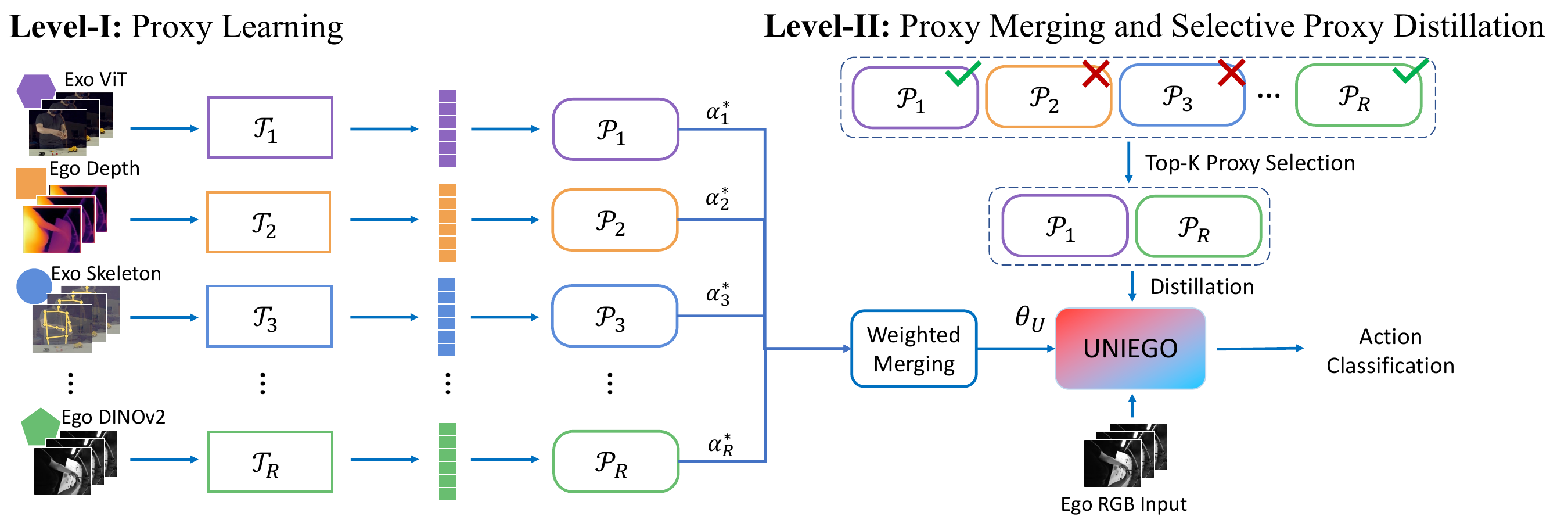}
  \caption{\textbf{Overview of \modelname.} \modelname~learns a unified egocentric encoder through a two-level proxy-mediated distillation framework. In Level-I (left), heterogeneous teachers spanning viewpoints, modalities, and foundation representations independently supervise egocentric proxy models, converting diverse teacher signals into a homogeneous proxy space. In Level-II (right), the proxy parameters are first merged to initialize the unified model, after which Selective Proxy Distillation (\stageIIabbr) performs sample-wise reliability filtering and distills only from proxies that are correct and confident. At inference, only the final \modelname~model operates using egocentric input.}
  \label{fig:architecture} \vspace{-5mm}
\end{figure}

\section{Method: Hierarchical Distillation Framework}
\label{sec:method}

In this section, we present \modelname, a hierarchical multi-teacher distillation framework for egocentric video understanding. Let $\mathcal{D}=\{(x_i^{e}, \{x_i^{r}\}_{r=1}^R, y_i)\}_{i=1}^{N}$ denote the training set, where $x_i^{e}$ is the egocentric video clip, $x_i^{r}$ denotes the input to the $r$-th teacher model $\mathcal{T}_r$, and $y_i \in \mathcal{Y}$ is the ground-truth action label. In practice, $x_i^{r}$ may coincide with $x_i^{e}$, correspond to its exocentric counterpart, or constitute an entirely different modality (\emph{e.g.}, skeleton or depth), each paired with its associated teacher $\mathcal{T}_r$. Each teacher thus encodes a distinct representation, modality, or viewpoint. The objective is to learn a unified egocentric encoder $f(\cdot)$ such that the resulting representation $f(x^{e})$ subsumes the complementary knowledge of all teachers $\{\mathcal{T}_r(x^{r})\}_{r=1}^{R}$ during training, while requiring \emph{only} the egocentric stream $x^{e}$ at inference.



\noindent\textbf{Overview.} \modelname~employs a hierarchical knowledge distillation strategy comprising two levels, as illustrated in Figure~\ref{fig:architecture}. In the \textit{first level}, we train a set of representation-specific \textbf{Proxy} models by independently distilling knowledge from each teacher $\mathcal{T}_r$. Each proxy specializes in transferring representation- or modality- or viewpoint-specific information from its corresponding teacher, while bridging the domain gap between heterogeneous teacher representations and the egocentric input space, converting diverse supervisory signals into a homogeneous egocentric proxy space. In the \textit{second level}, we perform Selective Proxy Distillation (SPD), which aggregates knowledge from all proxy models  into the final \modelname~model, initialized via a principled proxy merging strategy. The resulting \textbf{Unified Egocentric Model} effectively aggregates complementary knowledge from multiple teachers, with proxies acting as intermediaries for structured knowledge transfer. In the following subsections, we detail each component of \modelname's learning paradigm.


\subsection{Level-I: Proxy Learning}
In the first distillation level, we train $R$ proxy models $\{P_r\}_{r=1}^{R}$ by independently distilling feature-level knowledge from each teacher $\mathcal{T}_r$ into an egocentric student. 
Thus, for an egocentric video $x_i^e$ and its time-synchronized auxiliary inputs $\{x_i^r\}_{r=1}^{R}$ comprising RGB, depth, and skeleton data from both egocentric and exocentric viewpoints, we extract representations via $R$ teacher networks $\{\mathcal{T}_r\}_{r=1}^{R}$, each providing viewpoint- and modality-specific supervision signals.

All proxies share the same architecture but are trained with independent parameters. Since some teachers are foundation models that yield only feature embeddings rather than action logits, we adopt feature-level distillation throughout the entirety of this stage. Specifically, let $h_i, z_i = f(x_i^e)$ denote the feature embedding and action logits of the proxy student, and $h_i^r = \mathcal{T}_r(x_i^r)$ the teacher's feature embedding. Each proxy $P_r$ is optimized via:
\begin{equation}
    \mathcal{L}_{\text{I}}^{r} = \frac{1}{N} \sum_{i=1}^{N} \left( \lambda_{\text{I}}\, D_{\cos}(h_i, h_i^{r}) + \lambda_{\text{cls}}\, \operatorname{CE}(z_i, y_i) \right),
\end{equation}
where $D_{\cos}$ denotes cosine distance and $\operatorname{CE}$ is the cross-entropy loss.

Despite being induced by heterogeneous teachers, all proxies share the same egocentric architecture. This converts heterogeneous teacher supervision into a homogeneous egocentric proxy space $\{P_r\}_{r=1}^{R}$, partitioned as $\mathcal{P} = \mathcal{P}_{\text{ego}} \cup \mathcal{P}_{\text{exo}}$ according to viewpoint. Each proxy $P_r$ subsequently serves as a mediator in the second distillation level, reducing the modality and viewpoint gap between the original teachers and the final unified model.

\subsection{Level-II: Selective Proxy Distillation (SPD)}

Given the learned proxy set $\{P_r\}_{r=1}^{R}$, the goal of this level is to consolidate their complementary knowledge into a single unified egocentric model $f(\theta_U; \cdot)$. We first introduce a principled initialization of $\theta_U$ via proxy merging, followed by \textbf{SPD}, which selectively transfers knowledge from the most reliable proxies for each training sample.

\noindent\textbf{Proxy Merging Initialization.} We initialize $\theta_U$ as an optimally weighted combination of the proxy parameters. Let $\theta_r$ denote the parameters of proxy $P_r$. We solve for merging coefficients $\alpha^{*} \in \Delta^{R}$ that minimize the action classification loss over the training set:
\begin{equation}
    \theta_U \leftarrow \theta^{*}_{\text{merge}}, \quad \theta^{*}_{\text{merge}} = \sum_{r=1}^{R} \alpha_r^{*}\, \theta_r, \quad \alpha^{*} = \underset{\alpha \in \Delta^{R}}{\arg\min}\ \frac{1}{N}\sum_{i=1}^{N}\operatorname{CE}\!\left(f\!\left(\sum_{r=1}^{R} \alpha_r \theta_r;\, x_i^e\right), y_i\right),
\end{equation}
where $\Delta^{R} = \{\alpha \in \mathbb{R}^R_{\geq 0} \mid \sum_{r=1}^{R} \alpha_r = 1\}$ is the probability simplex and $f(\theta; x_i^e)$ denotes the model with parameters $\theta$ evaluated on $x_i^e$. This initialization places $\theta_U$ in a favourable region of the optimization landscape, providing a stable starting point for subsequent distillation.

\begin{proposition}[Proxy Merged Initialization as a Loss Upper Bound for \modelname]
Let $\mathcal{L}(\theta) = \frac{1}{N}\sum_{i=1}^{N} \operatorname{CE}(f(\theta; x_i^e), y_i)$ denote the classification loss. Assume $\mathcal{L}$ is convex in $\theta$ in a neighbourhood containing $\{\theta_r\}_{r=1}^{R}$. Then for any $\alpha \in \Delta^{R}$:
\begin{equation}
    \mathcal{L}\!\left(\sum_{r=1}^{R} \alpha_r \theta_r\right) \leq \sum_{r=1}^{R} \alpha_r\, \mathcal{L}(\theta_r),
\end{equation}
and consequently the optimally merged initialization satisfies: $\mathcal{L}(\theta^{*}_{\text{merge}}) \leq \min_{r}\, \mathcal{L}(\theta_r).$
\end{proposition}
The first inequality follows directly from Jensen's inequality~\cite{boyd2004convex} applied to the convex loss $\mathcal{L}$. The second follows by noting that $\alpha^{*}$ is chosen to minimize $\mathcal{L}(\sum_r \alpha_r \theta_r)$, and the degenerate solution $\alpha_r = 1$ for any single $r$ is feasible in $\Delta^{R}$, so the optimum is no worse than the best individual proxy.
This establishes that $\theta^{*}_{\text{merge}}$ achieves lower classification loss than any individual proxy under local convexity, placing \modelname~in a flatter, better-generalizing region of the loss landscape~\cite{izmailov2018averaging, frankle2020linear}. This favorable initialization reduces the optimization burden of SPD, as distillation begins from a point that already encodes the consensus of all $R$ proxy representations rather than the bias of any single one.

\noindent\textbf{Proxy Selection.} Rather than distilling from all proxies uniformly, SPD selects a reliable subset $\mathcal{S}_i \subseteq \{1, \dots, R\}$ for each sample $x_i^e$. We adopt a correctness-filtered small-loss criterion: a proxy $P_r$ is considered a reliable candidate only if it correctly predicts the action class, i.e., $\hat{y}_i^r = y_i$, where $\hat{y}_i^r = \arg\max_c (z_i^r)_c$. The candidate set is thus:
\begin{equation}
    \mathcal{C}_i = \{r \in \{1,\dots,R\} \mid \hat{y}_i^{r} = y_i\}.
\end{equation}
Among candidates, proxy reliability is quantified by the cross-entropy loss $s_i^r = \operatorname{CE}(z_i^{r}, y_i)$, where a lower loss indicates higher predictive confidence. When $|\mathcal{C}_i| > 0$, the top-$k$ proxies with the lowest $s_i^r$ are selected to form $\mathcal{S}_i$. When $\mathcal{C}_i = \emptyset$, we set $\mathcal{S}_i = \emptyset$ and skip distillation entirely for sample $i$, preventing the unified model from absorbing erroneous supervision.

\noindent\textbf{Selective Proxy Distillation.} For samples where $\mathcal{S}_i \neq \emptyset$, SPD transfers knowledge from the selected proxies to \modelname. Since all proxies are trained for action classification, they provide both feature embeddings and action logits. We therefore combine feature-level and logit-level distillation. Let $h_i^U, z_i^U = f(\theta_U; x_i^e)$ and $h_i^{P_r}, z_i^{P_r} = f(\theta_r; x_i^e)$ denote the feature embeddings and action logits of \modelname~and the $r$-th proxy $P_r$, respectively. The per-sample SPD loss is:
\begin{equation}
    \mathcal{L}_{\text{distill}}^{U}(i) = \frac{\beta_{\text{feat}}}{|\mathcal{S}_i|} \sum_{j \in \mathcal{S}_i} D_{\cos}(h_i^{U}, h_i^{P_j}) + \frac{\beta_{\text{logit}}}{|\mathcal{S}_i|} \sum_{j \in \mathcal{S}_i} D_{\text{KL}}\!\left(\sigma(z_i^{P_j}) \| \sigma(z_i^{U})\right),
\end{equation}
where $D_{\cos}$ and $D_{\text{KL}}$ denote cosine distance and KL divergence, $\sigma(\cdot)$ is the softmax function, and $\beta_{\text{feat}}, \beta_{\text{logit}}$ are loss weights. The total Level-II objective is:
\begin{equation}
    \mathcal{L}_{\text{II}} = \frac{1}{N} \sum_{i=1}^{N} \left( \mathbb{I}(|\mathcal{S}_i| > 0)\, \mathcal{L}_{\text{distill}}^{U}(i) + \beta_{\text{cls}}\operatorname{CE}(z_i^{U}, y_i) \right),
\end{equation}
where $\beta_{\text{cls}}$ is the classification loss weight and $\mathbb{I}(\cdot)$ is the indicator function. During inference, only \modelname~with egocentric input $x^e$ is required:
\begin{equation}
    \hat{y} = \arg\max_{c}\ z_c^{U}.
\end{equation}
No teacher network, exocentric stream, or proxy is needed at test time.

%% file: sec/4_experiments.tex
\section{Experiments}
\label{sec:exp}

We evaluate \modelname~on three egocentric datasets across three video understanding tasks: action recognition, video retrieval, and action segmentation. In all experiments, \modelname~requires only egocentric RGB video at inference.


\subsection{Experimental Setting}

\begin{wraptable}{o}{0.5\textwidth}
\centering
\caption{\textbf{Teacher pool for Level-I proxy learning.}}
\label{tab:teacher_modalities}
\resizebox{\linewidth}{!}{
\begin{tabular}{c c c c}
\toprule
\textbf{Modality} & \textbf{Viewpoint} & \textbf{Teacher Model} & \textbf{Feature Dim.} \\
\midrule
\multirow{2}{*}{RGB}     & Ego & DINOv2~\cite{oquab2024dinov2}        & 1024 \\
   & Ego & SigLIP~\cite{zhai2023siglipv1}        & 1152 \\
Depth    & Ego & DepthAnything~\cite{yang2024depthanythingv2} & 1024 \\
\midrule
\multirow{4}{*}{RGB}      & Exo & TimeSformer~\cite{timesformer}   & 768  \\
   & Exo & DINOv2~\cite{oquab2024dinov2}        & 1024 \\
   & Exo & Sk-Ego~\cite{reilly2025egoexo-v2-skego}        & 512  \\
   & Exo & SigLIP~\cite{zhai2023siglipv1}        & 1152 \\
Skeleton & Exo & ST-GCN~\cite{stgcn}        & 256  \\
Depth    & Exo & DepthAnything~\cite{yang2024depthanythingv2} & 1024 \\
\bottomrule
\end{tabular}}
\end{wraptable}
\textbf{Datasets.} We evaluate on three publicly available ego-exo datasets: \textbf{EgoExo-Fitness}~\cite{egoexofitness}, \textbf{Assembly101}~\cite{Sener2022Assembly101AL}, and \textbf{EgoExo4D}~\cite{EgoExo4D}. For EgoExo-Fitness and EgoExo4D, we follow the official evaluation splits~\cite{egoexofitness, EgoExo4D}. For Assembly101, we pair egocentric videos from the helmet-mounted \textit{ego04} camera with exocentric videos from the frontal \textit{exo03} camera, yielding 46,202 training and 15,307 test samples across 24 action classes. Further dataset details are provided in the Appendix.

\textbf{Implementation Details.}
We use TimeSformer~\cite{timesformer} as the egocentric student backbone for \modelname~ $f(\cdot)$ unless stated, which takes $8$ frames sampled at  $224 \times 224 $ and produces a 768-dimensional video representation. During training, \modelname~learns from multiple teacher representations $\mathcal{T}_r$ which are detailed in Table~\ref{tab:teacher_modalities}. We apply a linear projection layer to align all teacher representations into the 768-dimensional student space  to handle the dimensional mismatch between the student ($h_i$) and various teacher features ($h_i^r)$. All proxies are trained, following TimeSformer~\cite{timesformer} default configuration, for 15 epochs using SGD with a base learning rate of 0.005, momentum of 0.9, and weight decay of $1\times10^{-4}$. We scale the learning rate by factors of 0.1 and 0.01 at epochs 11 and 14, respectively. All models are trained with a total batch size of 8 distributed across 4 NVIDIA RTX A5000 GPUs.

In Level-I, we train 9 proxies across 6 teacher architectures and ego-exo viewpoints (Table~\ref{tab:teacher_modalities}), with loss weights $\lambda_{\text{I}}=5$ and $\lambda_{\text{cls}}=1$. In Level-II, the proxy merging coefficients $\alpha^{*}$ are optimized on the training set using Adam~\cite{adam_optimizer} for 2 epochs with a learning rate of $0.02$ and weight decay of $0.01$. For proxy selection, we set $K=1$ for EgoExo-Fitness and Assembly101, and $K=2$ for EgoExo4D (ablated in Table~\ref{tab:exp:ablation-adaptivedist-strategies}). The Level-II loss weights are $\beta_{\text{feat}}=5$, $\beta_{\text{cls}}=1$ throughout, and $\beta_{\text{logit}}=1$ for EgoExo-Fitness and EgoExo4D and $5$ for Assembly101, with KL divergence temperature $\tau=1$. All hyperparameters are selected on the respective validation sets.

\noindent

\begin{table*}[h!]
    \centering
    \setlength{\tabcolsep}{6pt}
    \renewcommand{\arraystretch}{1.2}

    \caption{Comparison with state-of-the-art methods on three egocentric action recognition datasets (EgoExo-Fitness, Assembly101, EgoExo4D). Dist. indicates whether the model performs distillation and Acc. indicates Top-1 accuracy.}
    \label{tab:exp:sota-all}

    \begin{subtable}[t]{0.3\textwidth}
        \centering
        \caption{\textbf{EgoExo-Fitness.}}
        \resizebox{\linewidth}{!}{
        \begin{tabular}{lcc}
             \hline
             \textbf{Method} & \textbf{Dist.} & \textbf{Acc.} \\
             \hline
             \multicolumn{3}{c}{\cellcolor{gray!20}\textit{Exocentric inference}} \\
             TimeSformer~\cite{timesformer} & \xmark& 88.9 \\
             ST-GCN~\cite{stgcn} & \xmark & 87.5 \\
             \hline
             \multicolumn{3}{c}{\cellcolor{gray!20}\textit{Egocentric inference}} \\
             I3D~\cite{i3d} & \xmark & 74.7 \\
             EgoVLP~\cite{kevin2022egovlp} & \xmark & 74.7 \\
             ViFi-CLIP~\cite{vificlip} & \xmark & 81.8 \\
             \scalebox{1.45}{$\pi$}-ViT~\cite{pivit} & \checkmark & 80.1 \\
             TimeSformer~\cite{timesformer} & \xmark & 80.3 \\
             Multiteacher Dist.~\cite{timesformer} & \checkmark & 81.5 \\
             \rowcolor{neuripspurple}\textbf{\modelname~(Ours)} & \checkmark & \textbf{84.7} \\
             \hline
        \end{tabular}}
        \label{tab:exp:sota-egoexofitness}
    \end{subtable}
    \hfill
    \begin{subtable}[t]{0.31\textwidth}
        \centering
        \caption{\textbf{Assembly101.}}
        \resizebox{\linewidth}{!}{
        \begin{tabular}{lcc}
             \hline
             \textbf{Method} & \textbf{Dist.} & \textbf{Acc.} \\
             \hline
             \multicolumn{3}{c}{\cellcolor{gray!20}\textit{Exocentric inference}} \\
             TimeSformer~\cite{timesformer} & \xmark & 62.7 \\
             ST-GCN~\cite{stgcn} & \xmark & 46.2 \\
             \hline
             \multicolumn{3}{c}{\cellcolor{gray!20}\textit{Egocentric inference}} \\
             TSM+TA~\cite{lin2019tsm, tempagg} & \xmark & 40.5 \\
             ViFi-CLIP~\cite{vificlip} & \xmark & 46.6 \\
             \scalebox{1.45}{$\pi$}-ViT~\cite{pivit} & \checkmark & 47.8 \\
             TimeSformer~\cite{timesformer} & \xmark & 47.6 \\
             Multiteacher Dist.~\cite{timesformer} & \checkmark & 48.2 \\
             \rowcolor{neuripspurple}\textbf{\modelname~(Ours)} & \checkmark & \textbf{50.7} \\
             \hline
        \end{tabular}}
        \label{tab:exp:sota-assembly101}
    \end{subtable}
    \hfill
    \begin{subtable}[t]{0.36\textwidth}
        \centering
        \caption{\textbf{EgoExo4D.}}
        \resizebox{\linewidth}{!}{
        \begin{tabular}{lcc}
             \hline
             \textbf{Method} & \textbf{Dist.} & \textbf{Acc.} \\
             \hline
             \multicolumn{3}{c}{\cellcolor{gray!20}\textit{Exocentric inference}} \\
             TimeSformer~\cite{timesformer} & \xmark & 26.0 \\
             ST-GCN~\cite{stgcn} & \xmark & 42.9 \\
             \hline
             \multicolumn{3}{c}{\cellcolor{gray!20}\textit{Egocentric inference}} \\
             VI Encoder~\cite{oord2018representation} & \checkmark & 40.3 \\
             EgoVLPv2~\cite{egovlpv2} & \checkmark & 39.1 \\
             Ego-Exo MAE~\cite{li2021_egoexo-transfer} & \checkmark & 37.2 \\
             Viewpoint Distillation~\cite{knowledge_distillation_hinton2015} & \checkmark & 38.2 \\
             TimeSformer~\cite{timesformer} & \xmark & 39.9 \\
             Multiteacher Dist.~\cite{timesformer} & \checkmark & 40.6 \\
             \rowcolor{neuripspurple}\textbf{\modelname~(Ours)} & \checkmark & \textbf{41.1} \\
             \hline
        \end{tabular}}
        \label{tab:exp:sota-egoexo4d}
    \end{subtable}
\end{table*}

\begin{table}[h!]
    \begin{minipage}{0.49\linewidth}
        \centering
        \caption{Performance on Video Retrieval}
        \resizebox{\linewidth}{!}{
        \begin{tabular}{c|cc|cc|cc}
             \hline
             \multirow{2}{*}{\textbf{Method}} 
             & \multicolumn{2}{c|}{\textbf{EgoExo-Fitness}} 
             & \multicolumn{2}{c|}{\textbf{Assembly101}} 
             & \multicolumn{2}{c}{\textbf{EgoExo4D}} \\
             & \textbf{mAP} & \textbf{R@1} 
             & \textbf{mAP} & \textbf{R@1} 
             & \textbf{mAP} & \textbf{R@1} \\
             \hline
             Timesformer & 0.474 & 0.712 & 0.226 & 0.410 & 0.167 & 0.326 \\
             Multiteacher Dist. & 0.486 & 0.720 & 0.228 & 0.413 & 0.178 & 0.331 \\
             \rowcolor{neuripspurple}\textbf{\modelname~(Ours)} & \textbf{0.543} & \textbf{0.748} & \textbf{0.253} & \textbf{0.424} & \textbf{0.182} & \textbf{0.340} \\
             \hline
        \end{tabular}}
        \label{tab:exp:retrieval}
    \end{minipage}
    \hfill
    \begin{minipage}{0.49\linewidth}
        \centering
        \caption{Performance on Egocentric Temporal Action Segmentation on Assembly101.}
        \resizebox{\textwidth}{!}{
        \begin{tabular}{c|ccccc}
             \hline
             \multirow{2}{*}{\textbf{\shortstack{Feature Backbone \\ (Method)}}}
             & \multirow{2}{*}{\textbf{F1@10}}
             & \multirow{2}{*}{\textbf{F1@25}}
             & \multirow{2}{*}{\textbf{F1@50}}
             & \multirow{2}{*}{\textbf{Edit}}
             & \multirow{2}{*}{\textbf{Acc}} \\
             &  & & & &  \\
             \hline
             TimeSformer (Ego only) & 16.2 & 14.1 & 10.4 & 18.7 & 34.4 \\
             TimeSformer (Multiteacher Dist.) & 15.3 & 13.2 & 9.8 & 18.4 & 34.2\\
             \rowcolor{neuripspurple}\textbf{\modelname~(Ours)} & \textbf{19.6} & \textbf{16.9} & \textbf{12.3} & \textbf{19.4} & \textbf{34.7} \\
             \hline
        \end{tabular}}
        \label{tab:exp:temporal-action-segmentation}
    \end{minipage}\vspace{-0.2in}
\end{table}

\subsection{Comparison with State-of-the-Art}

Table~\ref{tab:exp:sota-all} compares \modelname~with recent methods across three egocentric action recognition benchmarks under two inference protocols: \textit{exocentric inference}, where exocentric streams are available at test time, and \textit{egocentric inference}, where only egocentric video is provided. The former serves as a privileged-view upper bound~\cite{egoexofitness, Sener2022Assembly101AL}, as exocentric cameras in these datasets capture global body pose and scene layout that are largely occluded in the egocentric field of view. All baselines except ST-GCN~\cite{stgcn} utilize RGB input at inference.

\modelname~consistently outperforms the TimeSformer backbone by \textcolor{codegreen}{+4.4\%}, \textcolor{codegreen}{+3.1\%}, and \textcolor{codegreen}{+1.2\%} on EgoExo-Fitness, Assembly101, and EgoExo4D respectively, demonstrating the benefit of consolidating diverse multi-teacher supervision. Against the strongest distillation baseline $\pi$-ViT~\cite{pivit}, \modelname~achieves gains of \textcolor{codegreen}{+4.6\%} and \textcolor{codegreen}{+2.9\%} on EgoExo-Fitness and Assembly101. Furthermore, \modelname~consistently surpasses naive multi-teacher distillation~\cite{ranzinger2024radio, shang2024theia}, confirming that proxy-mediated hierarchical distillation is essential for reconciling heterogeneous teacher representations into a unified egocentric encoder.

The largest gains are observed on EgoExo-Fitness, where exocentric proxies $\mathcal{P}_{\text{exo}}$ are particularly strong as many actions in this dataset involve full-body motion that is inherently occluded from the egocentric viewpoint, making exocentric supervision especially informative. Conversely, on EgoExo4D, exocentric proxies are weaker, as evidenced by the baseline TimeSformer achieving only 26.0\% under exocentric inference, below its egocentric counterpart. Nevertheless, \modelname~remains robust to this proxy inconsistency across viewpoints, as the adaptive selection mechanism in \stageIIabbr~suppresses unreliable proxies and routes supervision from the most discriminative sources available. Across all three benchmarks, \modelname~achieves state-of-the-art action recognition performance under egocentric inference.

\begin{wraptable}{r}{0.35\textwidth}\vspace{-0.2in}
        \centering
        \caption{\textbf{Backbone robustness.} \modelname\ improves across diverse egocentric backbones.}
        \vspace{-0.15cm}
        \resizebox{\linewidth}{!}{
        \begin{tabular}{ccc}
             \hline
             \multirow{2}{*}{\textbf{Backbone}} & \multirow{2}{*}{\textbf{Method}} &
             \multirow{2}{*}{\textbf{EEF}} \\
             &  & \\
             \hline
             \multirow{3}{*}{TimeSformer~\cite{timesformer}} & baseline & 80.3 \\
              & Multiteacher Dist. & 81.5 \\
              & \textbf{\modelname~(Ours)} & \textbf{84.7} \\
             \hline
             \multirow{3}{*}{Uniformer-S~\cite{UniFormer}} & baseline & 68.4 \\
              & Multiteacher Dist. & 69.0 \\
              & \textbf{\modelname~(Ours)} & \textbf{73.5} \\
             \hline
              \multirow{3}{*}{ViFi-CLIP~\cite{vificlip}} & baseline  & 81.8 \\
              & Multiteacher Dist.& 81.7 \\
              & \textbf{\modelname~(Ours)} & \textbf{83.8} \\
             \hline
        \end{tabular}}
        \label{tab:exp:ablation-robustness}
\end{wraptable}\vspace{-0.05in}
\noindent\textbf{Generalization Across Backbone Architectures.} We verify that \modelname~is architecture-agnostic by replacing the proxy and unified model backbone with UniFormer-S~\cite{UniFormer} and ViFi-CLIP~\cite{vificlip} (Table~\ref{tab:exp:ablation-robustness}). Across both alternatives, our hierarchical distillation framework consistently outperforms naive multi-teacher distillation, confirming that the gains of \modelname~are not tied to a specific video encoder. Notably, UniFormer-S, a compact 22M model yields significant classification improvements, demonstrating that our framework is equally effective for learning efficient unified egocentric representations which is crucial for deployment in resource-constrained egocentric applications.

\noindent\textbf{Video Retrieval \& Temporal Action Segmentation.} Table~\ref{tab:exp:retrieval} evaluates \modelname~on video retrieval, performed by extracting features from action recognition trained backbones and computing their pairwise similarity across the test set. On EgoExo-Fitness, naive multi-teacher distillation yields only a marginal improvement of $+0.012$ mAP over the baseline, whereas SPD achieves a substantially larger gain of $+0.057$ mAP. This trend remains consistent across datasets, confirming that proxy-mediated distillation yields highly discriminative egocentric representations.

For temporal action segmentation on Assembly101, we extract features from three egocentric TimeSformer backbones: i) trained from scratch, ii) naive multi-teacher distillation, and iii) \modelname, and feed them into a fixed temporal encoder~\cite{sinha2025mstemba}. We evaluate the performance via F1 score, Edit distance, and frame accuracy. As shown in Table~\ref{tab:exp:temporal-action-segmentation}, \modelname~features yield the strongest performance across all metrics. Notably, naive multi-teacher distillation degrades performance on all metrics relative to the scratch baseline, confirming that directly distilling heterogeneous teachers disrupts fine-grained temporal representations. In contrast, \modelname's hierarchical distillation preserves the local temporal structure essential for frame-wise discrimination and boundary-sensitive segmentation.

\begin{table}[]
    \centering
    \setlength{\tabcolsep}{6pt}
    \renewcommand{\arraystretch}{1.2}

    \begin{minipage}{0.325\linewidth}
        \centering
        \caption{Ablation of components of \modelname.}
        \resizebox{\linewidth}{!}{
        \begin{tabular}{ccccc}
             \hline
             \multirow{2}{*}{\textbf{\shortstack{Proxy\\Learning}}} &
             \multirow{2}{*}{\textbf{\shortstack{Proxy\\Merging}}} &
             \multirow{2}{*}{\textbf{SPD}} &
             \multirow{2}{*}{\textbf{EEF}} &
             \multirow{2}{*}{\textbf{A101}} \\
              &  &  &  &  \\
             \hline
             \xmark & \xmark & \xmark & 80.3 & 47.6 \\
             \checkmark & \xmark & \xmark & 82.1 & 48.7 \\
             \checkmark & \xmark & \checkmark & 82.3 & 48.9 \\
             \checkmark & \checkmark & \xmark & 81.4 & 48.3 \\
             \rowcolor{neuripspurple} \checkmark & \checkmark & \checkmark & \textbf{84.7} & \textbf{50.7} \\
             \hline
        \end{tabular}}
        \label{tab:exp:ablation-components}
    \end{minipage}
    \hfill
    \begin{minipage}{0.325\linewidth}
        \centering
        \caption{Alternative proxy merging strategies.}
        \resizebox{\linewidth}{!}{
        \begin{tabular}{ccc}
             \hline
             \textbf{Merging Strategy} & \textbf{EEF} & \textbf{A101} \\
             \hline
             Best Proxy & 83.7 & 50.6\\
             Average & 83.4 & 50.2 \\
             Layer-level & 84.2 & 50.6 \\
             Parameter-level & 83.6 & 50.4 \\
             \rowcolor{neuripspurple}\textbf{Proxy Merging (Ours)} & \textbf{84.7} & \textbf{50.7} \\
             \hline
        \end{tabular}}
        \label{tab:exp:ablation-merging-strategies}
    \end{minipage}
    \hfill
    \begin{minipage}{0.31\linewidth}
        \centering
        \caption{Strategies for Selective Proxy distillation.}
        \resizebox{\linewidth}{!}{
        \begin{tabular}{ccc}
             \hline
             \textbf{Distillation Strategy} & \textbf{EEF} & \textbf{A101} \\
             \hline
             No Distillation & 80.3 & 47.6 \\
             All Proxies & 82.8 & 49.9 \\
             Top-1(Ego) + Top-1(Exo) & 83.2 & 49.5 \\
             Top-3 & 83.2 & 50.0 \\
             Top-2 & 84.3 & 50.1 \\
             \rowcolor{neuripspurple} Top-1 & \textbf{84.7} & \textbf{50.7} \\
             \hline
        \end{tabular}}
        \label{tab:exp:ablation-adaptivedist-strategies}
    \end{minipage}\vspace{-0.25in}
\end{table}

\subsection{Ablation Studies and Model Diagnosis}
We perform all the ablations and diagnosis of \modelname~on the EgoExo-Fitness (EEF) and Assembly101 (A101) datasets.

\noindent\textbf{Effect of Each Component.} Table~\ref{tab:exp:ablation-components} ablates the three key components of \modelname: Proxy Learning, Proxy Merging Initialization, and SPD. Starting from the egocentric-only baseline, adding Level-I proxy learning followed by simultaneous distillation from all proxies yields gains of \textcolor{codegreen}{+1.8\%} and \textcolor{codegreen}{+1.1\%} on EEF and A101, respectively, highlighting the role of proxies as mediators. Then, replacing the simultaneous distillation in level-II with SPD further improves performance by \textcolor{codegreen}{+0.2\%} on both datasets, demonstrating that instance-adaptive proxy selection yields more reliable supervision than distilling from all proxies indiscriminately. We also observe that proxy merging initialization provides a strong starting point for SPD, as the merged model already achieves higher accuracy than the baseline. Finally, prepending proxy merging initialization before SPD achieves 84.7\% and 50.7\% on EEF and A101 resulting in overall improvements of \textcolor{codegreen}{+4.4\%} and \textcolor{codegreen}{+3.1\%} over the baseline. These results confirm the contribution of each component in our hierarchical distillation framework.

\noindent\textbf{Alternative proxy merging strategies}
Table~\ref{tab:exp:ablation-merging-strategies} compares different strategies for initializing \modelname~before SPD. \textit{Best Proxy} uses the strongest individual proxy as the initialization, while \textit{Average} uniformly averages the weights of all trained proxies. We also compare against fine-grained learnable merging strategies: \textit{Layer-level} learns separate merging weights for each layer, and \textit{Parameter-level} learns separate merging weights for individual parameters. Although fine-grained merging offers greater flexibility, it does not improve performance in practice. Both layer-level and parameter-level merging outperform uniform averaging yet fall short of our global merging strategy, suggesting that a globally consistent combination of proxy parameters is more effective than local merging at the level of individual layers or parameter groups, likely due to the importance of maintaining parameter consistency across the full model.


\noindent\textbf{Alternative SPD strategies.}
In Table~\ref{tab:exp:ablation-adaptivedist-strategies}, we investigate various selective distillation strategies following the proxy merging initialization. 
SPD outperforms distillation from all proxies simultaneously, confirming that selective supervision mitigates conflicting gradients across homogeneous proxies. Enforcing viewpoint diversity via a \textit{Top-1(Ego) + Top-1(Exo)} selection strategy degrades performance, particularly on Assembly101, indicating that global proxy selection is preferable, i.e., not all training samples contain discriminative information from every viewpoint, and forcing viewpoint-balanced selection introduces noisy supervision into \modelname. Among $K \in \{1, 2, 3\}$, all top-$K$ variants outperform naive proxy distillation, demonstrating the robustness of the selection mechanism, with $K=1$ yielding the best overall performance.



\section{Does \modelname~Mitigate Representational Gap and Conflicting Gradients?}
The answer is ``\textit{Yes}".
We provide a model analysis to confirm that our hierarchical distillation framework addresses the two core failure modes of naive multi-teacher distillation: representational gap and conflicting gradients.

\begin{figure}[t]
  \centering
  \begin{minipage}{0.30\linewidth}
    \centering
    \includegraphics[width=\linewidth, height=0.85\linewidth, keepaspectratio]{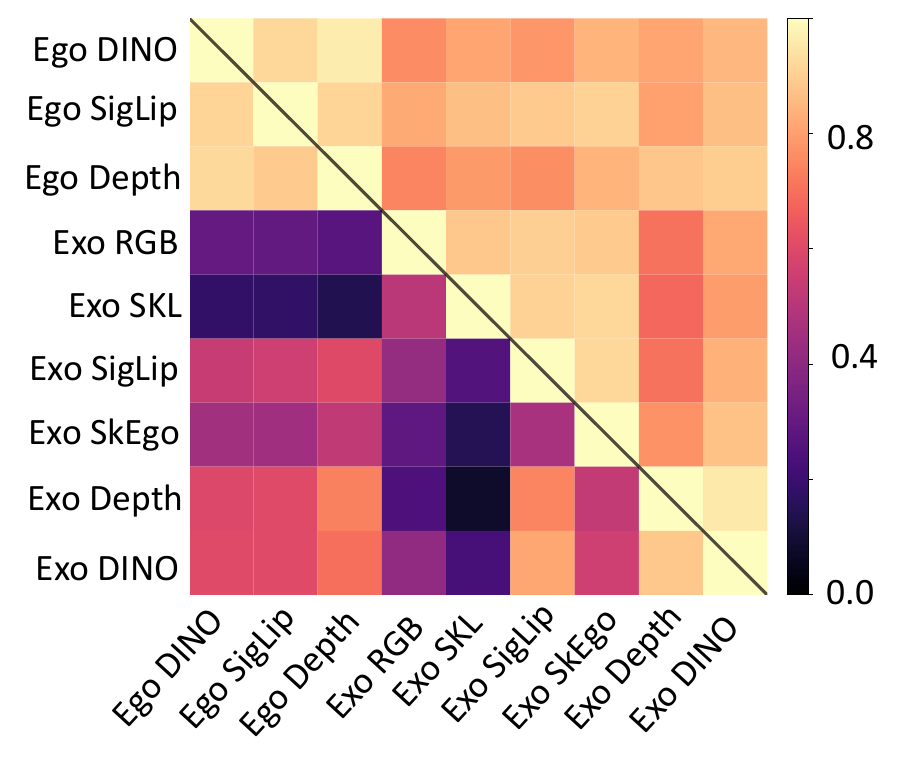}
    \caption{Teacher vs Proxy Centered Kernel Alignment scores} %
    \label{fig:representation_gap}
  \end{minipage}
  \hfill
  \begin{minipage}{0.32\linewidth}
    \centering
    \includegraphics[width=\linewidth]{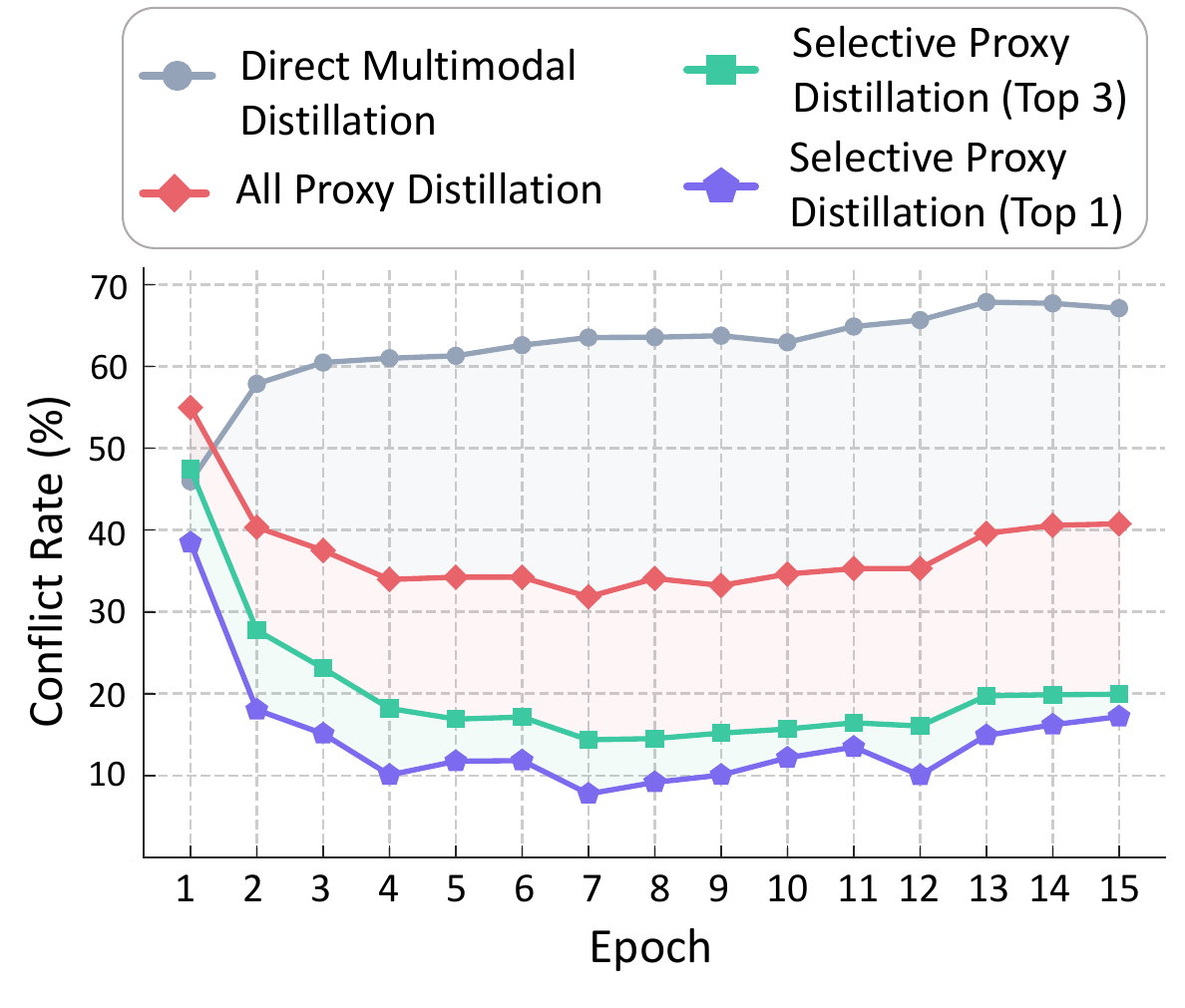}
    \caption{Gradient Conflict Rate across distillation strategies}
    \label{fig:gradient_conflict_rate}
  \end{minipage}
  \hfill
  \begin{minipage}{0.34\linewidth}
    \centering
    \includegraphics[width=\linewidth]{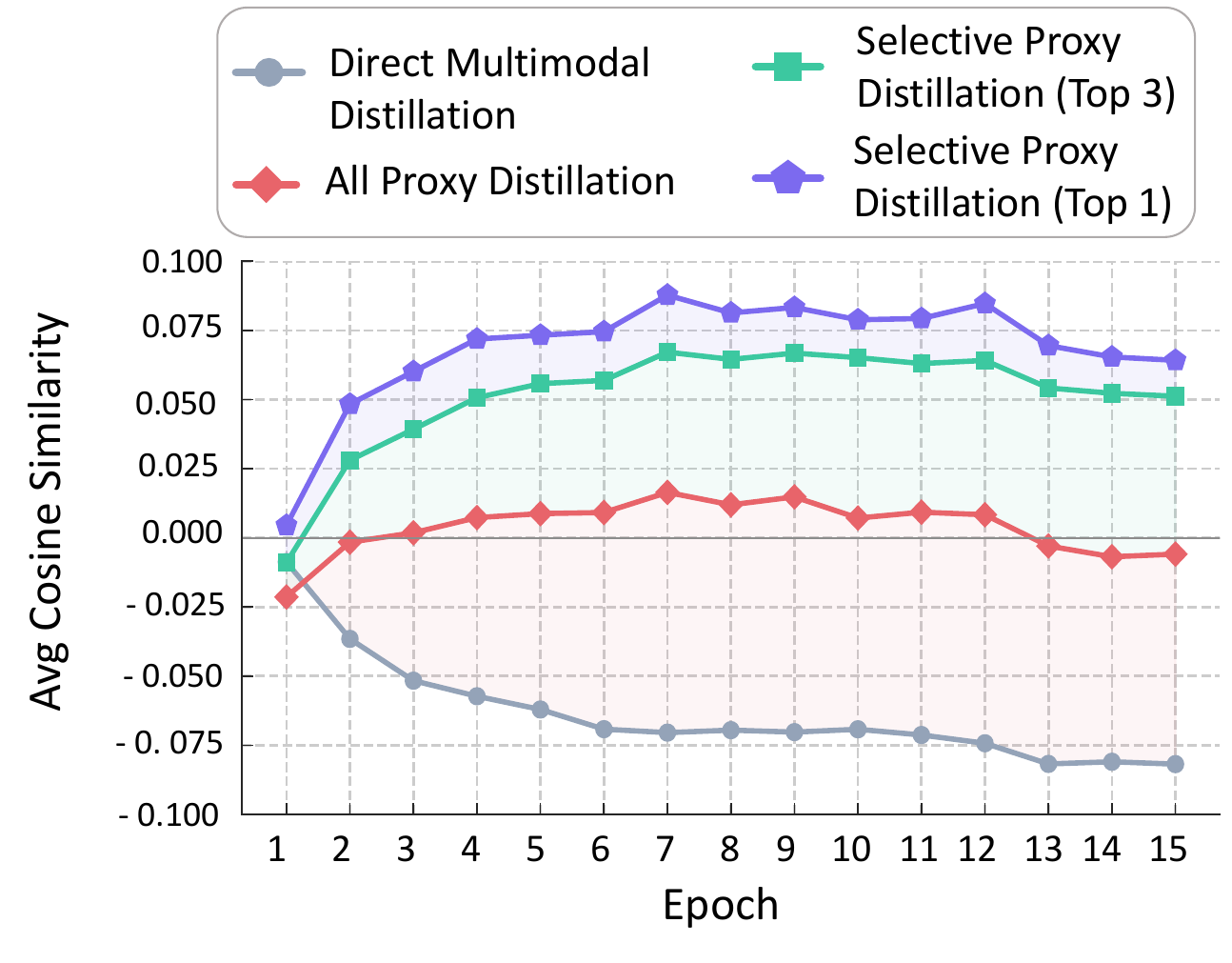}
    \caption{Cosine Similarity among teacher and proxy gradients} %
    \label{fig:gradient_cosine_similarity}
  \end{minipage}
  \vspace{-5mm}
\end{figure}

\noindent\textbf{Representational Gap.} To quantify the representational gap among the $R$ teachers, we visualize their pairwise linear Centered Kernel Alignment (CKA)~\cite{kornblith2019cka-similarity} in Figure~\ref{fig:representation_gap}. The lower triangle displays teacher-pair similarities and the upper triangle displays proxy-pair similarities on EgoExo-Fitness. The substantially higher pairwise CKA among proxies than among teachers confirms that Level-I proxy learning projects heterogeneous teacher representations into a homogeneous egocentric embedding space, directly alleviating the representational gap that impedes naive multi-teacher distillation.

\noindent\textbf{Conflicting Gradients.} Figures~\ref{fig:gradient_conflict_rate} and~\ref{fig:gradient_cosine_similarity} report the gradient conflict rate which is defined as the fraction of distillation gradients that oppose the classification gradient, i.e., $\cos(\nabla_{\text{cls}}, \nabla_{\text{kd}}) < 0$, and the average cosine similarity across training epochs, respectively. Direct multi-teacher distillation exhibits severe gradient interference, confirming that heterogeneous teacher supervision actively opposes the primary task objective. Distilling from all proxies simultaneously reduces the conflict rate to $42$--$50\%$ and shifts the cosine similarity toward zero, indicating partial mitigation. SPD further suppresses conflicts by routing supervision exclusively through reliable proxies, yielding consistently cooperative distillation and classification gradients. Together, these results confirm that proxy learning and SPD collectively transform a conflicting multi-teacher optimization landscape into a coherent one.

%% file: sec/5_conclusion.tex
\section{Conclusion}
\label{sec:conc}

We presented \textbf{\modelname}, a unified egocentric encoder that consolidates diverse perceptual knowledge across viewpoints, modalities, and foundation model representations into a single model. The core of \modelname~lies a hierarchical distillation framework in which representation-specific proxies serve as structured mediators, dissolving the representational gap between heterogeneous teachers before selective, reliability-guided distillation assembles their collective knowledge into a coherent unified representation. The result is an egocentric encoder that sees further, knows more, and generalizes better without  requiring more than a single egocentric camera at inference. 

While we hope \modelname~inspires broader exploration into systematically harvesting and reconciling diverse supervisory signals for richer egocentric perception, the current framework has an important limitation. Specifically, \modelname~relies on a small-loss criterion for proxy selection, a heuristic that, while effective, does not exploit the full potential of the proxy pool. A learned selection mechanism, one that dynamically weighs proxy reliability as a function of both the input and the training state could yield richer and more adaptive supervision. Designing such an adaptive proxy selection strategy is non-trivial, and we leave this as a promising direction for future work.



\section*{Acknowledgements}
This work was supported in part by the National Science Foundation (IIS-2245652) and the University of North Carolina at Charlotte. Computational resources were provided by the NSF National AI Research Resource Pilot (NAIRR240338) and NCShare.

%% file: sec/6_supplementary.tex
\label{sec:appendix}

\section{Overview}
The appendix is categorized into the following parts: \\
\begin{itemize}
    \item Section \ref{sec:data_description}: Detailed Data Description
    \item Section \ref{sec:proxy performance}: Proxy Performances 
    \item Section \ref{sec:analysis}: Analysis of Proxy Selection
    \item Section \ref{sec:proxy merging}: Analysis of Proxy Merging
    \item Section \ref{sec:action analysis}: Action Analysis
\end{itemize}

\section{Detailed Dataset Description} \label{sec:data_description}
\textbf{EgoExo-Fitness}~\cite{egoexofitness} is a full-body action understanding dataset containing synchronized egocentric and exocentric videos of fitness activities. It consists of 12 action categories and approximately 32 hours of video. Following the official split, we use 3,522 training samples and 912 test samples, with action sequences ranging from 10 to 30 seconds in duration. 

\textbf{Assembly101}~\cite{Sener2022Assembly101AL} is a large-scale procedural activity dataset comprising 167 hours of video, recording subjects performing object assembly tasks from multiple viewpoints. In our experiments, we pair the egocentric videos from the helmet-mounted \textit{ego04} camera with the corresponding exocentric videos in front \textit{exo03} camera. The resulting split contains 46,202 training samples and 15,307 test samples over 24 action classes. 

\textbf{EgoExo4D}~\cite{EgoExo4D} is a large-scale egocentric-exocentric video dataset covering a diverse range of skilled human activities. For each sample, we pair the egocentric video with the exocentric view annotated as the best view by human annotators. The dataset split used in our experiments contains 30,660 training samples and 9,356 test samples across 665 action classes.

\section{Proxy Performances} \label{sec:proxy performance}

\begin{table}[h]
\centering
\caption{Performance of the base ego model and different ego/exo proxy models across EgoExo-Fitness (EEF), Assembly101 (A101), and Ego-Exo4D (EE4D).}
\resizebox{0.6\linewidth}{!}{
\begin{tabular}{c c c c c c}
\toprule
\textbf{Modality} & \textbf{Viewpoint} & \textbf{Model} & 
\textbf{EEF} & \textbf{A101} & \textbf{EE4D} \\
\midrule

\multicolumn{6}{c}{\textbf{\underline{Base Ego Model}}} \\
RGB~\cite{timesformer} & Ego & TimeSformer & 80.3 & 47.6 & 39.9 \\

\midrule
\multicolumn{6}{c}{\textbf{\underline{Ego Proxies} ($\mathcal{P}_{ego}$)}} \\

Depth~\cite{yang2024depthanythingv2} & Ego & DepthAnything & 80.3 & 48.2 & 40.1 \\
DINOv2~\cite{oquab2024dinov2} & Ego & DINOv2 & 80.9 & 48.4 & 40.2 \\
SigLip~\cite{zhai2023siglipv1} & Ego & SigLip & 80.7 & 48.4 & 40.6 \\

\midrule
\multicolumn{6}{c}{\textbf{\underline{Exo Proxies} ($\mathcal{P}_{exo}$)}} \\

RGB~\cite{timesformer} & Exo & TimeSformer & 81.9 & 49.5 & 40.0 \\
Skeleton~\cite{stgcn} & Exo & ST-GCN & 81.9 & 48.4 & 40.3 \\
Depth~\cite{yang2024depthanythingv2} & Exo & DepthAnything & 80.3 & 48.5 & 39.9 \\
DINOv2~\cite{oquab2024dinov2} & Exo & DINOv2 & 80.7 & 48.2 & 39.5 \\
Sk-Ego~\cite{reilly2025egoexo-v2-skego} & Exo & Sk-Ego & 81.9 & 48.3 & 40.2 \\
SigLip~\cite{zhai2023siglipv1} & Exo & SigLip & 80.3 & 48.0 & 40.6 \\

\bottomrule
\end{tabular}
}
\label{tab:proxy_performance}
\vspace{-4mm}
\end{table}

In Table~\ref{tab:proxy_performance}, we report the action classification performance of all proxies ($\mathcal{P}_r$) after Level-I distillation. We observe that not all proxies outperform the base ego model, motivating the need for proxy selection. Nevertheless, the final performance of \modelname~surpasses that of all individual proxies, corroborating the effectiveness of our proposed hierarchical distillation framework.

\section{Analysis of Proxy Selection.} \label{sec:analysis}
\begin{figure}[t]
  \centering
  \scalebox{0.8}{
  \includegraphics[width=0.5\linewidth]{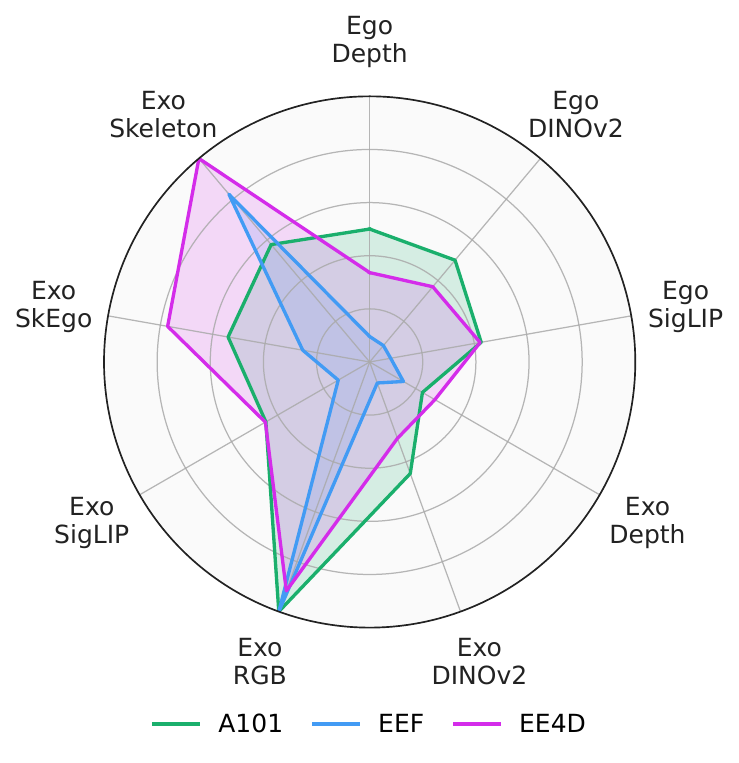}}
  \caption{\textbf{Selective Proxy Distillation Statistics.}}
  \label{fig:proxy_selection}
  \vspace{-5mm}
\end{figure}

In Figure~\ref{fig:proxy_selection}, we analyze the selection frequency of each proxy during the Level-II selective proxy distillation (SPD).
First, every proxy is selected to some extent across all benchmarks, indicating that SPD does not collapse to a single teacher.
Unsurprisingly, we find that the routing distributions are not consistent across benchmarks, and that proxy selection is strongly correlated with the distinct dataset characteristics.
On \textbf{EgoExo-Fitness}, the SPD primarily selects \textit{Exo Skeleton} and \textit{Exo RGB}. This is consistent with the nature of the dataset, which consists largely of full-body fitness motions: exocentric and skeleton-based proxies provide supervision for body pose and motion patterns that are often only partially visible from a head-mounted camera.
In \textbf{Assembly101}, the selection is more diverse, assigning a larger weight to egocentric proxies. This suggests that, compared to EgoExo-Fitness, more actions in Assembly101 rely on egocentric cues such as local hand-object interactions and subtle manipulation dynamics.
On \textbf{EgoExo4D}, there is a notable increase in the selection of the \textit{Exo SkEgo} proxy. We attribute this to the high-resolution, spatially localized nature of the benchmark: action-relevant regions can occupy only a small part of the full frame, and the SkEgo crop provides a more focused view of these regions.

\section{Analysis of Proxy Merging} \label{sec:proxy merging}

\begin{table}[t]
\centering
\caption{\textbf{Effect of different proxy initialization strategies before and after SPD on EgoExo-Fitness (EEF) and Assembly101 (A101).}}
\label{tab:proxy_merge}
\vspace{1mm}
\resizebox{0.65\linewidth}{!}{
\begin{tabular}{l c c c c}
\toprule
\multirow{2}{*}{\textbf{Method}} & 
\multicolumn{2}{c}{\textbf{EEF}} & 
\multicolumn{2}{c}{\textbf{A101}} \\
\cmidrule(lr){2-3} \cmidrule(lr){4-5}
& \textbf{Before SPD} & \textbf{After SPD} & 
\textbf{Before SPD} & \textbf{After SPD} \\
\midrule

Best Proxy       & 81.9 & 83.7 & 49.5 & 50.6 \\
Average          & 79.9 & 83.4 & 42.9 & 50.2 \\
Layer-level      & 80.9 & 84.2 & 43.5 & 50.6 \\
Parameter-level  & 79.8 & 83.6 & 43.4 & 50.4 \\
Model-level      & 81.4 & 84.7 & 48.3 & 50.7 \\

\bottomrule
\end{tabular}
}
\vspace{-4mm}
\end{table}

Table~\ref{tab:proxy_merge} compares action classification performance of \modelname~under different proxy merging initialization strategies, before and after SPD, on EgoExo-Fitness and Assembly101. Across all initialization strategies, SPD consistently improves performance, confirming the importance of the second distillation level. Notably, the best-performing initialization prior to SPD, i.e., the single Best Proxy, does not yield the strongest post-SPD accuracy. Instead, the model soup, i.e., the learned convex combination of all proxy parameters, provides the most effective initialization for SPD. This suggests that the advantage of proxy merging lies not in its standalone classification performance but in its ability to place \modelname~in a well-conditioned region of the loss landscape that is maximally amenable to subsequent distillation.

\section{Action Analysis} \label{sec:action analysis}

\begin{table}[t]
\centering
\caption{Per-class top-1 proxy selection rates on EEF (12 classes)}
\label{tab:proxy_eef}
\resizebox{\textwidth}{!}{%
\footnotesize
\begin{tabular}{lcccccccccc}
\toprule
class\_name & none & exo\_rgb & exo\_skl & exo\_siglip & ego\_siglip & exo\_skego & ego\_depth & exo\_depth & exo\_dino & ego\_dino \\
\midrule
Kneeling pushing-ups            & 0.053 & 0.139 & 0.172 & 0.053 & 0.053 & 0.139 & 0.013 & \textbf{0.179} & 0.106 & 0.093 \\
Push-ups                        & 0.000 & \textbf{0.500} & 0.400 & 0.000 & 0.025 & 0.042 & 0.008 & 0.017 & 0.000 & 0.008 \\
Kneeling Torso Twist            & 0.046 & \textbf{0.431} & 0.218 & 0.030 & 0.005 & 0.056 & 0.117 & 0.041 & 0.036 & 0.020 \\
Knee Raise \& Abd. Contract     & 0.055 & 0.290 & \textbf{0.428} & 0.014 & 0.028 & 0.048 & 0.055 & 0.034 & 0.021 & 0.028 \\
Shoulder Bridge                 & 0.000 & \textbf{0.987} & 0.013 & 0.000 & 0.000 & 0.000 & 0.000 & 0.000 & 0.000 & 0.000 \\
Sit-ups                         & 0.057 & 0.172 & \textbf{0.261} & 0.070 & 0.013 & 0.108 & 0.076 & 0.146 & 0.045 & 0.051 \\
Leg Reverse Lunge               & 0.038 & 0.106 & \textbf{0.477} & 0.068 & 0.068 & 0.114 & 0.023 & 0.045 & 0.053 & 0.008 \\
Leg Lunge With Knee Lift        & 0.118 & \textbf{0.300} & 0.291 & 0.073 & 0.036 & 0.109 & 0.036 & 0.018 & 0.009 & 0.009 \\
Sumo Squat                      & 0.112 & \textbf{0.388} & 0.233 & 0.017 & 0.017 & 0.198 & 0.009 & 0.009 & 0.000 & 0.017 \\
Jumping Jacks                   & 0.042 & 0.242 & \textbf{0.347} & 0.126 & 0.016 & 0.153 & 0.016 & 0.005 & 0.021 & 0.032 \\
High Knee                       & 0.054 & 0.264 & \textbf{0.271} & 0.085 & 0.078 & 0.078 & 0.008 & 0.054 & 0.054 & 0.054 \\
Clap Jacks                      & 0.011 & 0.345 & \textbf{0.414} & 0.029 & 0.023 & 0.063 & 0.029 & 0.057 & 0.011 & 0.017 \\
\bottomrule
\end{tabular}%
}
\label{tab:action_anlysis_EEF}
\end{table}

\begin{table}[t]
\centering
\caption{Per-class top-1 proxy selection rates on A101 (24 classes).}
\label{tab:proxy_a101}
\resizebox{\textwidth}{!}{%
\footnotesize
\begin{tabular}{lcccccccccc}
\toprule
class\_name & none & exo\_rgb & exo\_skl & exo\_siglip & ego\_siglip & exo\_skego & ego\_depth & exo\_depth & exo\_dino & ego\_dino \\
\midrule
pick up               & 0.049 & \textbf{0.256} & 0.125 & 0.089 & 0.079 & 0.104 & 0.079 & 0.042 & 0.076 & 0.102 \\
put down              & 0.051 & \textbf{0.204} & 0.122 & 0.107 & 0.106 & 0.103 & 0.091 & 0.045 & 0.076 & 0.094 \\
inspect               & 0.186 & \textbf{0.136} & 0.108 & 0.080 & 0.066 & 0.092 & 0.085 & 0.052 & 0.091 & 0.105 \\
rotate                & 0.307 & \textbf{0.123} & 0.085 & 0.056 & 0.049 & 0.090 & 0.088 & 0.037 & 0.083 & 0.082 \\
unscrew               & 0.042 & \textbf{0.196} & 0.124 & 0.100 & 0.100 & 0.123 & 0.124 & 0.044 & 0.064 & 0.085 \\
position              & 0.107 & 0.139 & 0.081 & 0.075 & 0.086 & \textbf{0.143} & 0.125 & 0.048 & 0.088 & 0.108 \\
screw                 & 0.048 & \textbf{0.250} & 0.078 & 0.099 & 0.087 & 0.088 & 0.128 & 0.042 & 0.087 & 0.091 \\
remove                & 0.187 & 0.125 & \textbf{0.138} & 0.064 & 0.067 & 0.098 & 0.090 & 0.044 & 0.085 & 0.102 \\
position screw on     & 0.134 & \textbf{0.141} & 0.086 & 0.104 & 0.084 & 0.088 & 0.111 & 0.039 & 0.107 & 0.106 \\
remove screw from     & 0.201 & 0.115 & 0.117 & 0.092 & 0.068 & 0.071 & 0.083 & 0.043 & \textbf{0.141} & 0.071 \\
pass                  & 0.165 & \textbf{0.142} & 0.089 & 0.059 & 0.067 & 0.123 & 0.086 & 0.040 & 0.130 & 0.098 \\
tilt up               & 0.291 & \textbf{0.143} & 0.041 & 0.058 & 0.082 & 0.057 & 0.102 & 0.042 & 0.087 & 0.097 \\
attempt to position   & 0.413 & 0.084 & \textbf{0.097} & 0.036 & 0.025 & 0.088 & 0.089 & 0.035 & 0.048 & 0.085 \\
push                  & 0.336 & 0.063 & 0.098 & 0.076 & 0.096 & \textbf{0.102} & 0.061 & 0.054 & 0.078 & 0.036 \\
tilt down             & 0.224 & \textbf{0.179} & 0.116 & 0.078 & 0.068 & 0.108 & 0.046 & 0.041 & 0.079 & 0.059 \\
pull                  & 0.370 & 0.082 & \textbf{0.123} & 0.042 & 0.062 & 0.071 & 0.076 & 0.027 & 0.120 & 0.029 \\
attempt to remove     & 0.595 & 0.067 & 0.031 & 0.029 & 0.012 & 0.084 & 0.061 & 0.006 & 0.027 & \textbf{0.088} \\
attempt to pick up    & 0.754 & 0.040 & 0.022 & 0.025 & 0.020 & \textbf{0.047} & 0.025 & 0.017 & 0.017 & 0.032 \\
clap                  & 0.007 & \textbf{0.883} & 0.047 & 0.010 & 0.005 & 0.015 & 0.005 & 0.007 & 0.012 & 0.007 \\
attempt to unscrew    & 0.515 & 0.095 & 0.075 & 0.037 & 0.050 & 0.025 & 0.079 & 0.008 & 0.017 & \textbf{0.100} \\
attempt to put down   & 0.988 & \textbf{0.012} & 0.000 & 0.000 & 0.000 & 0.000 & 0.000 & 0.000 & 0.000 & 0.000 \\
spin                  & 0.452 & \textbf{0.170} & 0.064 & 0.053 & 0.027 & 0.059 & 0.037 & 0.027 & 0.032 & 0.080 \\
attempt to screw      & 0.951 & \textbf{0.019} & 0.010 & 0.000 & 0.000 & 0.010 & 0.000 & 0.000 & 0.010 & 0.000 \\
shake                 & 0.783 & \textbf{0.075} & 0.000 & 0.025 & 0.000 & 0.017 & 0.008 & 0.000 & 0.042 & 0.050 \\
\bottomrule
\end{tabular}%
}
\label{tab:action_anlysis_A101}
\end{table}

As shown in Table~\ref{tab:action_anlysis_EEF} and Table~\ref{tab:action_anlysis_A101}, \texttt{exo\_rgb} is the most frequently selected proxy on both datasets,
with a sample-weighted selection rate of $0.349$ on EEF and $0.182$ on A101,
followed by \texttt{exo\_skl}. However, the class-level separation between the
top-ranked and second-ranked proxies differs substantially across datasets.
The average per-class gap between the top-1 and top-2 proxies is $0.186$ on EEF,
but only $0.069$ on A101. This suggests that EEF classes often have a more
clearly preferred proxy, whereas A101 exhibits a flatter proxy distribution,
where several proxies can have similar selection rates for the same class.

Interestingly, \texttt{exo\_depth} shows a highly class-specific behavior.
Although it has the lowest aggregate selection rate among all proxies
($0.042$ on A101 and $0.052$ on EEF), it is the most frequently selected proxy
for \textit{Kneeling push-ups} on EEF, with a selection rate of $0.179$,
exceeding the second-ranked proxy, \texttt{exo\_rgb}, at $0.139$.
This indicates that depth can provide useful supervision for specific classes
where geometric cues, such as body-to-floor distance, are discriminative.
However, its low aggregate selection rate also suggests that this benefit is
not broadly shared across classes. 

The behavior of ego-side proxies also differs across the two datasets.
On EEF, all ego-side proxies have relatively low selection rates, around
$0.03$, suggesting that third-person supervision is more reliable for large
whole-body fitness motions. In contrast, on A101, \texttt{ego\_dino} and
\texttt{ego\_depth} reach selection rates of approximately $0.091$ and rank
among the middle group of proxies. Each of them is also the most frequently
selected proxy for at least one \textit{attempt-to-*} class. This difference
is consistent with the task characteristics of A101: fine-grained assembly
actions are often hand- and object-centric, and egocentric observations may
capture local manipulation details that are less visible from exocentric
views. 



%% file: dominick_ref.bib
@inproceedings{zhai2023siglipv1,
      title={Sigmoid Loss for Language Image Pre-Training}, 
      author={Zhai, Xiaohua and Mustafa, Basil and Kolesnikov, Alexander and Beyer, Lucas},
      booktitle={Proceedings of the IEEE/CVF International Conference on Computer Vision},
      year={2023},
      pages={}, 
      organization={IEEE}
}

@misc{reilly2025egoexo-v2-skego,
      title={From My View to Yours: Ego-Augmented Learning in Large Vision Language Models for Understanding Exocentric Daily Living Activities}, 
      author={Reilly, Dominick and Govind, Manish Kumar and Xue, Le and Das, Srijan},
      year={2025},
      eprint={2501.05711},
      archivePrefix={arXiv}
}

@inproceedings{yang2024depthanythingv2,
      title={Depth Anything V2},
      author={Yang, Lihe and Kang, Bingyi and Huang, Zilong and Zhao, Zhen and Xu, Xiaogang and Feng, Jiashi and Zhao, Hengshuang},
      booktitle={Advances in Neural Information Processing Systems},
      year={2024}
}

@inproceedings{grauman2022ego4d,
  title        = {Ego4D: Around the World in 3,000 Hours of Egocentric Video},
  author       = {Grauman, Kristen and Westbury, Andrew and Byrne, Eugene and Chavis, Zachary and Furnari, Antonino and Girdhar, Rohit and Hamburger, Jackson and Jiang, Hao and Liu, Miao and Liu, Xingyu and Martin, Miguel and Nagarajan, Tushar and Radosavovic, Ilija and Ramakrishnan, Santhosh Kumar and Ryan, Fiona and Sharma, Jayant and Wray, Michael and Xu, Mengmeng and Xu, Eric Zhongcong and Zhao, Chen and Bansal, Siddhant and Batra, Dhruv and Cartillier, Vincent and Crane, Sean and Do, Tien and Doulaty, Morrie and Erapalli, Akshay and Feichtenhofer, Christoph and Fragomeni, Adriano and Fu, Qichen and Gebreselasie, Abrham and Gonzalez, Cristina and Hillis, James and Huang, Xuhua and Huang, Yifei and Jia, Wenqi and Khoo, Weslie and Kolar, Jachym and Kottur, Satwik and Kumar, Anurag and Landini, Federico and Li, Chao and Li, Yanghao and Li, Zhenqiang and Mangalam, Karttikeya and Modhugu, Raghava and Munro, Jonathan and Murrell, Tullie and Nishiyasu, Takumi and Price, Will and Ruiz Puentes, Paola and Ramazanova, Merey and Sari, Leda and Somasundaram, Kiran and Southerland, Audrey and Sugano, Yusuke and Tao, Ruijie and Vo, Minh and Wang, Yuchen and Wu, Xindi and Yagi, Takuma and Zhao, Ziwei and Zhu, Yunyi and Arbelaez, Pablo and Crandall, David and Damen, Dima and Farinella, Giovanni Maria and Fuegen, Christian and Ghanem, Bernard and Ithapu, Vamsi Krishna and Jawahar, C. V. and Joo, Hanbyul and Kitani, Kris and Li, Haizhou and Newcombe, Richard and Oliva, Aude and Park, Hyun Soo and Rehg, James M. and Sato, Yoichi and Shi, Jianbo and Shou, Mike Zheng and Torralba, Antonio and Torresani, Lorenzo and Yan, Mingfei and Malik, Jitendra},
  booktitle    = {Proceedings of the IEEE/CVF Conference on Computer Vision and Pattern Recognition},
  year         = {2022},
  pages        = {18995--19012}
}

@inproceedings{gong2023mmgego4d,
  title     = {MMG-Ego4D: Multi-Modal Generalization in Egocentric Action Recognition},
  author    = {Xinyu Gong and Sreyas Mohan and Naina Dhingra and Jean-Charles Bazin and Yilei Li and Zhangyang Wang and Rakesh Ranjan},
  booktitle = {Proceedings of the IEEE/CVF Conference on Computer Vision and Pattern Recognition (CVPR)},
  year      = {2023}
}

@inproceedings{xu2025egodtm,
  title     = {EgoDTM: Towards 3D-Aware Egocentric Video-Language Pretraining},
  author    = {Boshen Xu and Yuting Mei and Xinbi Liu and Sipeng Zheng and Qin Jin},
  booktitle = {Advances in Neural Information Processing Systems (NeurIPS)},
  year      = {2025}
}

@article{tan2023egodistill,
  title   = {EgoDistill: Egocentric Head Motion Distillation for Efficient Video Understanding},
  author  = {Shuhan Tan and Tushar Nagarajan and Kristen Grauman},
  journal = {arXiv preprint arXiv:2301.02217},
  year    = {2023}
}

@inproceedings{radevski2023multimodal,
  title     = {Multimodal Distillation for Egocentric Action Recognition},
  author    = {Gorjan Radevski and Dusan Grujicic and Marie-Francine Moens and Matthew Blaschko and Tinne Tuytelaars},
  booktitle = {Proceedings of the IEEE/CVF International Conference on Computer Vision (ICCV)},
  year      = {2023}
}


%% file: ref.bib
@String(CVPR= {Conference on Computer Vision and Pattern Recognition})

@String(ICCV= {International Conference on Computer Vision
})

@String(ECCV= {European Conference on Computer Vision})

@String(ICPR = {Int. Conf. Pattern Recog.})

@String(ICASSP=	{ICASSP})

@String(ICLR = {International Conference on Learning Representations})

@String(AAAI = {AAAI})

@inproceedings{das2020vpn,
  title        = {Vpn: Learning video-pose embedding for activities of daily living},
  author       = {Das, Srijan and Sharma, Saurav and Dai, Rui and Bremond, Francois and Thonnat, Monique},
  booktitle    = {European Conference on Computer Vision},
  pages        = {72--90},
  year         = {2020},
  organization = {Springer}
}

@inproceedings{izmailov2018averaging,
  title={Averaging Weights Leads to Wider Optima and Better Generalization},
  author={Izmailov, Pavel and Podoprikhin, Dmitrii and Garipov, Timur and Vetrov, Dmitry and Wilson, Andrew Gordon},
  booktitle={Uncertainty in Artificial Intelligence (UAI)},
  year={2018}
}

@inproceedings{frankle2020linear,
  title={Linear Mode Connectivity and the Lottery Ticket Hypothesis},
  author={Frankle, Jonathan and Dziugaite, Gintare Karolina and Roy, Daniel M and Carlin, Michael},
  booktitle={International Conference on Machine Learning (ICML)},
  year={2020}
}

@book{boyd2004convex,
  title={Convex Optimization},
  author={Boyd, Stephen and Vandenberghe, Lieven},
  publisher={Cambridge University Press},
  year={2004}
}

@article{OpenPose,
  author  = {Zhe {Cao} and Gines {Hidalgo Martinez} and Tomas {Simon} and Shih-En {Wei} and Yaser {Sheikh}},
  journal = {IEEE Transactions on Pattern Analysis and Machine Intelligence},
  title   = {OpenPose: Realtime Multi-Person 2D Pose Estimation using Part Affinity Fields},
  year    = {2019}
}

@inproceedings{i3d,
  title        = {Quo vadis, Action Recognition? A New Model and the Kinetics Dataset},
  author       = {Carreira, Joao and Zisserman, Andrew},
  booktitle    = CVPR,
  pages        = {4724--4733},
  year         = {2017},
  organization = {IEEE}
}

@inproceedings{ranzinger2024radio,
  title={Am-radio: Agglomerative vision foundation model reduce all domains into one},
  author={Ranzinger, Mike and Heinrich, Greg and Kautz, Jan and Molchanov, Pavlo},
  booktitle={Proceedings of the IEEE/CVF conference on computer vision and pattern recognition},
  year={2024}
}

@article{shang2024theia,
  title={Theia: Distilling diverse vision foundation models for robot learning},
  author={Shang, Jinghuan and Schmeckpeper, Karl and May, Brandon B and Minniti, Maria Vittoria and Kelestemur, Tarik and Watkins, David and Herlant, Laura},
  journal={arXiv preprint arXiv:2407.20179},
  year={2024}
}

@misc{ticon,
      title={TICON: A Slide-Level Tile Contextualizer for Histopathology Representation Learning}, 
      author={Varun Belagali and Saarthak Kapse and Pierre Marza and Srijan Das and Zilinghan Li and Sofiène Boutaj and Pushpak Pati and Srikar Yellapragada and Tarak Nath Nandi and Ravi K Madduri and Joel Saltz and Prateek Prasanna and Stergios Christodoulidis and Maria Vakalopoulou and Dimitris Samaras},
      year={2025},
      eprint={2512.21331},
      archivePrefix={arXiv},
      primaryClass={cs.CV},
      url={https://arxiv.org/abs/2512.21331}, 
}

@inproceedings{wimmer2026anyup,
    title={AnyUp: Universal Feature Upsampling},
    author={Wimmer, Thomas and Truong, Prune and Rakotosaona, Marie-Julie and Oechsle, Michael and Tombari, Federico and Schiele, Bernt and Lenssen, Jan Eric},
    booktitle={Proceedings of the International Conference on Learning Representations ({ICLR})},
    year={2026}
}

@inproceedings{
shi2023recon,
title={Recon: Reducing Conflicting Gradients from the Root for Multi-Task Learning},
author={Guangyuan Shi and Qimai Li and Wenlong Zhang and Jiaxin Chen and Xiao-Ming Wu},
booktitle={International Conference on Learning Representations (ICLR)},
year={2023},
url={https://openreview.net/forum?id=ivwZO-HnzG_}
}

@inproceedings{conflict_adverse_GD,
 author = {Liu, Bo and Liu, Xingchao and Jin, Xiaojie and Stone, Peter and Liu, Qiang},
 booktitle = {Advances in Neural Information Processing Systems},
 editor = {M. Ranzato and A. Beygelzimer and Y. Dauphin and P.S. Liang and J. Wortman Vaughan},
 pages = {18873--18885},
 publisher = {Curran Associates, Inc.},
 title = {Conflict-Averse Gradient Descent for Multi-task Learning},
 url = {https://proceedings.neurips.cc/paper/2021/file/9d27fdf2477ffbff837d73ef7ae23db9-Paper.pdf},
 volume = {34},
 year = {2021}
}

@article{sinha2025mstemba,
                title={MS-Temba: Multi-Scale Temporal Mamba for Efficient Temporal Action Detection},
                author={Sinha, Arkaprava and Raj, Monish Soundar and Wang, Pu and Helmy, Ahmed and Das, Srijan},
                journal={arXiv preprint arXiv:2501.06138},
                year={2025}
                }

@inproceedings{PCGRAD,
 author = {Yu, Tianhe and Kumar, Saurabh and Gupta, Abhishek and Levine, Sergey and Hausman, Karol and Finn, Chelsea},
 booktitle = {Advances in Neural Information Processing Systems},
 editor = {H. Larochelle and M. Ranzato and R. Hadsell and M. F. Balcan and H. Lin},
 pages = {5434--5445},
 publisher = {Curran Associates, Inc.},
 title = {Gradient Surgery for Multi-Task Learning},
 url = {https://proceedings.neurips.cc/paper/2020/file/3fe78a8acf5fda99de95303940a2420c-Paper.pdf},
 volume = {33},
 year = {2020}
}

@article{oord2018representation,
  title   = {Representation Learning with Contrastive Predictive Coding},
  author  = {Oord, Aaron van den and Li, Yazhe and Vinyals, Oriol},
  journal = {arXiv preprint arXiv:1807.03748},
  year    = {2018}
}

@inproceedings{stgcn,
  title     = {Spatial temporal graph convolutional networks for skeleton-based action recognition},
  author    = {Yan, Sijie and Xiong, Yuanjun and Lin, Dahua},
  booktitle = {Thirty-second AAAI conference on artificial intelligence},
  year      = {2018}
}

@article{sigurdsson2018charades-ego,
  title   = {Charades-ego: A large-scale dataset of paired third and first person videos},
  author  = {Sigurdsson, Gunnar A and Gupta, Abhinav and Schmid, Cordelia and Farhadi, Ali and Alahari, Karteek},
  journal = {arXiv preprint arXiv:1804.09626},
  year    = {2018}
}

@inproceedings{Damen2018EPICKITCHENS,
  title     = {Scaling Egocentric Vision: The EPIC-KITCHENS Dataset},
  author    = {Damen, Dima and Doughty, Hazel and Farinella, Giovanni Maria  and Fidler, Sanja and Furnari, Antonino and Kazakos, Evangelos and Moltisanti, Davide and Munro, Jonathan 
               and Perrett, Toby and Price, Will and Wray, Michael},
  booktitle = {European Conference on Computer Vision (ECCV)},
  year      = {2018}
}

@inproceedings{lin2019tsm,
  title     = {TSM: Temporal Shift Module for Efficient Video Understanding},
  author    = {Lin, Ji and Gan, Chuang and Han, Song},
  booktitle = {Proceedings of the IEEE International Conference on Computer Vision},
  year      = {2019}
}

@inproceedings{adam_optimizer,
  title={Adam: A Method for Stochastic Optimization},
  author={Kingma, Diederik P and Ba, Jimmy},
  booktitle={International Conference on Learning Representations (ICLR)},
  year={2015}
}

@inproceedings{1,
  title     = {Learning realistic human actions from movies},
  author    = {Ivan Laptev and Marcin Marszałek and Cordelia Schmid and Benjamin Rozenfeld},
  booktitle = {CVPR},
  year      = {2008}
}

@inproceedings{2,
  title     = {Recognizing human actions: A local SVM approach},
  author    = {Christian Schuldt and Ivan Laptev and Barbara Caputo},
  booktitle = {ICPR},
  year      = {2004}
}

@inproceedings{3,
  title     = {Action recognition with improved trajectories},
  author    = {Heng Wang and Cordelia Schmid},
  booktitle = {ICCV},
  year      = {2013}
}

@inproceedings{4,
  title     = {Space-time Interest Points},
  author    = {Ivan Laptev and Tony Lindeberg},
  booktitle = {ICCV},
  year      = {2003}
}

@inproceedings{pivit,
  author    = {Dominick Reilly and Srijan Das},
  title     = {Just Add $\pi$! Pose Induced Video Transformers for Understanding Activities of Daily Living},
  booktitle = {Proceedings of the IEEE/CVF Conference on Computer Vision and Pattern Recognition (CVPR)},
  month     = {June},
  year      = {2024}
}

@article{kevin2022egovlp,
  title={Egocentric Video-Language Pretraining},
  author={Lin, Kevin Qinghong and Wang, Alex Jinpeng and Soldan, Mattia and Wray, Michael and Yan, Rui and Xu, Eric Zhongcong and Gao, Difei and Tu, Rongcheng and Zhao, Wenzhe and Kong, Weijie and others},
  journal={arXiv preprint arXiv:2206.01670},
  year={2022}
}

@inproceedings{timesformer,
  author    = {Gedas Bertasius and Heng Wang and Lorenzo Torresani},
  title     = {Is Space-Time Attention All You Need for Video Understanding?},
  booktitle = {Proceedings of the International Conference on Machine Learning (ICML)},
  month     = {July},
  year      = {2021}
}

@article{vpn++,
  author  = {Das, Srijan and Dai, Rui and Yang, Di and Bremond, Francois},
  journal = {IEEE Transactions on Pattern Analysis and Machine Intelligence},
  title   = {VPN++: Rethinking Video-Pose embeddings for understanding Activities of Daily Living},
  year    = {2021},
  volume  = {},
  number  = {},
  pages   = {1-1},
}

@inproceedings{dino,
  title     = {Emerging properties in self-supervised vision transformers},
  author    = {Caron, Mathilde and Touvron, Hugo and Misra, Ishan and J{\'e}gou, Herv{\'e} and Mairal, Julien and Bojanowski, Piotr and Joulin, Armand},
  booktitle = {Proceedings of the IEEE/CVF International Conference on Computer Vision},
  pages     = {9650--9660},
  year      = {2021}
}

@inproceedings{videopose3d,
  title     = {3D human pose estimation in video with temporal convolutions and semi-supervised training},
  author    = {Pavllo, Dario and Feichtenhofer, Christoph and Grangier, David and Auli, Michael},
  booktitle = {Conference on Computer Vision and Pattern Recognition (CVPR)},
  year      = {2019}
}

@misc{knowledge_distillation_hinton2015,
  title         = {Distilling the Knowledge in a Neural Network},
  author        = {Geoffrey Hinton and Oriol Vinyals and Jeff Dean},
  year          = {2015},
  eprint        = {1503.02531},
  archiveprefix = {arXiv},
  primaryclass  = {stat.ML}
}

@article{clip_representation,
  author   = {Qiuhong {Ke} and Mohammed {Bennamoun} and Senjian {An} and Ferdous {Sohel} and Farid {Boussaid}},
  journal  = {IEEE Transactions on Image Processing},
  title    = {Learning Clip Representations for Skeleton-Based 3D Action Recognition},
  year     = {2018},
  volume   = {27},
  number   = {6},
  pages    = {2842-2855},
  doi      = {10.1109/TIP.2018.2812099},
  issn     = {1941-0042},
  month    = {June}
}

@inproceedings{vificlip,
  title     = {Finetuned CLIP models are efficient video learners},
  author    = {Rasheed, Hanoona and Khattak, Muhammad Uzair and Maaz, Muhammad and Khan, Salman and Khan, Fahad Shahbaz},
  booktitle = {The IEEE/CVF Conference on Computer Vision and Pattern Recognition},
  year      = {2023}
}

@article{ego_exo_survey,
title = {Egocentric and exocentric methods: A short survey},
journal = {Computer Vision and Image Understanding},
volume = {257},
pages = {104371},
year = {2025},
issn = {1077-3142},
author = {Anirudh Thatipelli and Shao-Yuan Lo and Amit K. Roy-Chowdhury},
}

@article{lavila,
  title={Learning Video Representations from Large Language Models},
  author={Yue Zhao and Ishan Misra and Philipp Krahenbuhl and Rohit Girdhar},
  journal={2023 IEEE/CVF Conference on Computer Vision and Pattern Recognition (CVPR)},
  year={2022},
  pages={6586-6597},
  url={https://api.semanticscholar.org/CorpusID:254408789}
}

@article{UniFormer,
  title={UniFormer: Unifying Convolution and Self-Attention for Visual Recognition},
  author={Kunchang Li and Yali Wang and Junhao Zhang and Peng Gao and Guanglu Song and Yu Liu and Hongsheng Li and Yu Jiao Qiao},
  journal={IEEE Transactions on Pattern Analysis and Machine Intelligence},
  year={2022},
  volume={45},
  pages={12581-12600},
  url={https://api.semanticscholar.org/CorpusID:246240170}
}

@InProceedings{xue2023_egoexo-ae2,
      title={Learning Fine-grained View-Invariant Representations from Unpaired Ego-Exo Videos via Temporal Alignment},
      author={Xue, Zihui and Grauman, Kristen},
      booktitle={Advances in Neural Information 
Processing Systems (NeurIPS)},
      year={2023}
}

@InProceedings{li2021_egoexo-transfer,
 title={Ego-Exo: Transferring Visual Representations from Third-person to First-person Videos},
 author={Li, Yanghao and Nagarajan, Tushar and Xiong, Bo and Grauman, Kristen},
 booktitle={Proceedings of the IEEE/CVF Conference on Computer Vision and Pattern Recognition (CVPR)},
 pages={10995--11005},
 year={2021}
}

@InProceedings{quattrocchi2024_synch-all-you-need,
 title={Synchronization is All You Need: Exocentric-to-Egocentric Transfer for Temporal Action Segmentation with Unlabeled Synchronized Video Pairs},
 author={Quattrocchi, Camillo and Furnari, Antonino and Di Mauro, Daniele and Giuffrida, Mario Valerio and Farinella, Giovanni Maria},
 booktitle={European Conference on Computer Vision (ECCV)},
 year={2024}
}

@article{EgoExo4D,
  title={Ego-Exo4D: Understanding Skilled Human Activity from First- and Third-Person Perspectives},
  author={Kristen Grauman and Andrew Westbury and Lorenzo Torresani and Kris Kitani and Jitendra Malik and Triantafyllos Afouras and Kumar Ashutosh and Vijay Baiyya and Siddhant Bansal and Bikram Boote and Eugene Byrne and Zachary Chavis and Joya Chen and Feng Cheng and Fu-Jen Chu and Sean Crane and Avijit Dasgupta and Jing Dong and Mar{\'i}a Escobar and Cristhian Forigua and Abrham Kahsay Gebreselasie and Sanjay Haresh and Jing Huang and Md Mohaiminul Islam and Suyog Dutt Jain and Rawal Khirodkar and Devansh Kukreja and Kevin J. Liang and Jia-Wei Liu and Sagnik Majumder and Yongsen Mao and Miguel Martin and Effrosyni Mavroudi and Tushar Nagarajan and Francesco Ragusa and Santhosh K. Ramakrishnan and Luigi Seminara and Arjun Somayazulu and Yale Song and Shan Su and Zihui Xue and Edward Zhang and Jinxu Zhang and {\'A}ngela Castillo and Changan Chen and Xinzhu Fu and Ryosuke Furuta and Cristina Gonz{\'a}lez and Prince Gupta and Jiabo Hu and Yifei Huang and Yiming Huang and Weslie Khoo and Anushk Kumar and Robert Kuo and Sach Lakhavani and Miao Liu and Romy Mi Luo and Zhengyi Luo and Brighid Meredith and Austin Miller and Oluwatumininu Oguntola and Xiaqing Pan and Penny Peng and Shraman Pramanick and Merey Ramazanova and Fiona Ryan and W. Shan and Kiran Somasundaram and Chenan Song and Audrey Southerland and Masatoshi Tateno and Huiyu Wang and Yuchen Wang and Takuma Yagi and Mingfei Yan and Xitong Yang and Ze Yu and Shengxin Zha and Chen Zhao and Ziwei Zhao and Zhifan Zhu and J. F. Zhuo and Pablo Arbel{\'a}ez and Gedas Bertasius and David J. Crandall and Dima Damen and Jakob Julian Engel and Giovanni Maria Farinella and Antonino Furnari and Bernard Ghanem and Judy Hoffman and C. V. Jawahar and Richard A. Newcombe and Hyun Soo Park and James M. Rehg and Yoichi Sato and Manolis Savva and Jianbo Shi and Mike Zheng Shou and Michael Wray},
  journal={2024 IEEE/CVF Conference on Computer Vision and Pattern Recognition (CVPR)},
  year={2023},
  pages={19383-19400},
  url={https://api.semanticscholar.org/CorpusID:265506384}
}

@inproceedings{egoexofitness,
  title={EgoExo-Fitness: towards egocentric and exocentric full-body action understanding},
  author={Li, Yuan-Ming and Huang, Wei-Jin and Wang, An-Lan and Zeng, Ling-An and Meng, Jing-Ke and Zheng, Wei-Shi},
  booktitle={European Conference on Computer Vision},
  pages={363--382},
  year={2024},
  organization={Springer}
}

@inproceedings{luo2025viewpoint,
      title={Viewpoint Rosetta Stone: Unlocking Unpaired Ego-Exo Videos for View-invariant Representation Learning},
      author={Luo, Mi and Xue, Zihui and Dimakis, Alex and Grauman, Kristen},
      booktitle={Proceedings of the IEEE/CVF Conference on Computer Vision and Pattern Recognition},
      year={2025}
    }

@misc{EMBED,
      title={Unlocking Exocentric Video-Language Data for Egocentric Video Representation Learning}, 
      author={Zi-Yi Dou and Xitong Yang and Tushar Nagarajan and Huiyu Wang and Jing Huang and Nanyun Peng and Kris Kitani and Fu-Jen Chu},
      year={2024},
      eprint={2408.03567},
      archivePrefix={arXiv},
      primaryClass={cs.CV},
      url={https://arxiv.org/abs/2408.03567}, 
}

@inproceedings{tempagg,
  title={Temporal aggregate representations for long-range video understanding},
  author={Sener, Fadime and Singhania, Dipika and Yao, Angela},
  booktitle={European conference on computer vision},
  pages={154--171},
  year={2020},
  organization={Springer}
}

@inproceedings{egovlpv2,
  title={Egovlpv2: Egocentric video-language pre-training with fusion in the backbone},
  author={Pramanick, Shraman and Song, Yale and Nag, Sayan and Lin, Kevin Qinghong and Shah, Hardik and Shou, Mike Zheng and Chellappa, Rama and Zhang, Pengchuan},
  booktitle={Proceedings of the IEEE/CVF International Conference on Computer Vision},
  pages={5285--5297},
  year={2023}
}

@article{oquab2024dinov2,
  title   = {DINOv2: Learning Robust Visual Features without Supervision},
  author  = {Maxime Oquab and Timoth{\'e}e Darcet and Theo Moutakanni and Huy V. Vo and Marc Szafraniec and Vasil Khalidov and Pierre Fernandez and Daniel Haziza and Francisco Massa and Alaaeldin El-Nouby and Russell Howes and Po-Yao Huang and Hu Xu and Vasu Sharma and Shang-Wen Li and Wojciech Galuba and Mike Rabbat and Mido Assran and Nicolas Ballas and Gabriel Synnaeve and Ishan Misra and Herv{\'e} J{\'e}gou and Julien Mairal and Patrick Labatut and Armand Joulin and Piotr Bojanowski},
  journal = {Transactions on Machine Learning Research (TMLR)},
  year    = {2024}
}

@inproceedings{kornblith2019cka-similarity,
  title     = {Similarity of Neural Network Representations Revisited},
  author    = {Simon Kornblith and Mohammad Norouzi and Honglak Lee and Geoffrey Hinton},
  booktitle = {Proceedings of the 36th International Conference on Machine Learning (ICML)},
  year      = {2019}
}

@inproceedings{zhang2022confidence,
  title     = {Confidence-Aware Multi-Teacher Knowledge Distillation},
  author    = {Hailin Zhang and Defang Chen and Can Wang},
  booktitle = {ICASSP 2022 - 2022 IEEE International Conference on Acoustics, Speech and Signal Processing (ICASSP)},
  year      = {2022}
}

@article{liu2021adaptive,
  title   = {Adaptive Multi-Teacher Multi-Level Knowledge Distillation},
  author  = {Yuang Liu and Wei Zhang and Jun Wang},
  journal = {Neurocomputing},
  year    = {2020}
}

@article{Sener2022Assembly101AL,
  title={Assembly101: A Large-Scale Multi-View Video Dataset for Understanding Procedural Activities},
  author={Fadime Sener and Dibyadip Chatterjee and Daniel Shelepov and Kun He and Dipika Singhania and Robert Wang and Angela Yao},
  journal={2022 IEEE/CVF Conference on Computer Vision and Pattern Recognition (CVPR)},
  year={2022},
  pages={21064-21074},
  url={https://api.semanticscholar.org/CorpusID:247762252}
}
